\documentclass[11pt]{article}

\usepackage[T1]{fontenc}
\usepackage[utf8]{inputenc}
\usepackage{lmodern}
\usepackage[margin=1in]{geometry}
\usepackage{amsmath,amssymb,amsthm,mathtools}
\usepackage{xcolor}
\usepackage{graphicx}
\usepackage{booktabs}
\usepackage{array}
\usepackage{enumitem}
\usepackage[font=small,labelfont=bf]{caption}
\usepackage{subcaption}
\usepackage[authoryear,round]{natbib}
\usepackage{microtype}
\usepackage[colorlinks=true,linkcolor=blue!50!black,citecolor=blue!50!black,urlcolor=blue!50!black]{hyperref}
\usepackage[capitalise,nameinlink]{cleveref}

\graphicspath{{figures/}}

\newcommand{\RR}{\mathbb{R}}
\newcommand{\CC}{\mathbb{C}}
\newcommand{\NN}{\mathbb{N}}

\newcommand{\EE}{\mathbb{E}}
\newcommand{\PP}{\mathbb{P}}

\newcommand{\TV}{\mathrm{TV}}
\newcommand{\Hel}{h}                                   
\newcommand{\chisq}{\chi^{2}}
\newcommand{\Wass}[1]{W_{#1}}                          

\newcommand{\Normal}{\mathcal{N}}                      
\newcommand{\dd}{\mathrm{d}}                           
\newcommand{\one}{\mathbf{1}}

\DeclareMathOperator{\supp}{supp}

\newcommand{\rbar}{\bar{r}}                 
\newcommand{\Mhn}{M}                        
\newcommand{\hnsys}[2]{(\mathrm{HN}_{#1}^{#2})} 
\newcommand{\Okc}{\mathcal{O}_{k,c_0}}      
\newcommand{\Ghat}{\widehat{G}_{n}}         

\newcommand{\hnHe}{\mathrm{He}}             
\newcommand{\hnsc}{S^{-}}                   
\newcommand{\hnphi}{\phi}                   
\newcommand{\hnmom}{\mathsf{m}}             
\newcommand{\hnE}{E}                        
\newcommand{\hne}{e}                        
\newcommand{\hnord}{\varkappa}              

\theoremstyle{plain}
\newtheorem{theorem}{Theorem}[section]
\newtheorem{proposition}[theorem]{Proposition}
\newtheorem{lemma}[theorem]{Lemma}
\newtheorem{corollary}[theorem]{Corollary}

\newtheorem{fact}[theorem]{Fact}
\theoremstyle{definition}
\newtheorem{definition}[theorem]{Definition}

\theoremstyle{remark}
\newtheorem{remark}[theorem]{Remark}
\numberwithin{equation}{section}

\crefname{theorem}{Theorem}{Theorems}
\crefname{proposition}{Proposition}{Propositions}
\crefname{lemma}{Lemma}{Lemmas}
\crefname{corollary}{Corollary}{Corollaries}
\crefname{conjecture}{Conjecture}{Conjectures}
\crefname{fact}{Fact}{Facts}
\crefname{definition}{Definition}{Definitions}
\crefname{assumption}{Assumption}{Assumptions}
\crefname{example}{Example}{Examples}
\crefname{remark}{Remark}{Remarks}
\crefname{section}{Section}{Sections}
\crefname{subsection}{Section}{Sections}
\crefname{appendix}{Appendix}{Appendices}
\crefname{figure}{Figure}{Figures}
\crefname{table}{Table}{Tables}
\crefformat{equation}{(#2#1#3)}
\Crefformat{equation}{(#2#1#3)}
\crefmultiformat{equation}{(#2#1#3)}{ and~(#2#1#3)}{, (#2#1#3)}{ and~(#2#1#3)}

\newcommand{\hnsection}{\section}
\newcommand{\hnsubsection}{\subsection}
\newcommand{\HNsystemdisplay}[2]{%
  \sum_{j=1}^{#1}\ \sum_{\substack{n_1,n_2\ge0\\ n_1+2n_2=\alpha}}
  \frac{c_j^{2}\,a_j^{n_1}\,b_j^{n_2}}{n_1!\,n_2!}=0,\qquad \alpha=1,\dots,#2 }

\newcommand{\HNmainstatement}{%
Let $\Mhn\ge2$ and $r\ge1$. The system $\hnsys{\Mhn}{r}$ has a real solution
$(c_j,a_j,b_j)_{j=1}^{\Mhn}$ with $c_j\neq0$ for all $j$ and $a_j\neq0$ for some $j$
if and only if $r\le 2\Mhn-1$. Moreover, every real solution of $\hnsys{\Mhn}{2\Mhn}$ with
$c_j\neq0$ for all $j$ satisfies $a_j=b_j=0$ for all $j$. Consequently, for location--scale
Gaussian mixtures with $k_0$ true and $k>k_0$ fitted components,
$\rbar(k-k_0)=2(k-k_0+1)$.}

\newcommand{\HNshortstatement}{%
the Ho--Nguyen order of over-fitted location--scale Gaussian mixtures is
$\rbar(k-k_0)=2(k-k_0+1)$ (real unknowns, all weights non-zero)}

\newcommand{\notetitle}{Solving the Ho--Nguyen Conjecture for Gaussian Mixtures: Applications to Mixtures of Experts}

\newcommand{\noteabstract}{%
Over-fitted Gaussian mixtures can contain many parameter configurations with nearly the same
density, making the order of density cancellation central to their parameter-learning rates.
Ho and Nguyen encoded these cancellations in a polynomial system and conjectured its exact
order as a function of cluster size. The key to resolving the conjecture is a common
positive-variance shift that turns the coefficient equations into Gaussian moment matching.
For every cluster size $\Mhn\ge2$, we prove that the real, nonzero-weight system has order
$2\Mhn$: a sharp sign-change bound forces rigidity at that order, while positive Gauss--Hermite
quadrature constructs a non-trivial solution through order $2\Mhn-1$.
Population-density calculations at 110-digit working precision recover the predicted leading
term for two to six components; finite multistart searches recover the same observed powers
for two to five components, with smaller finite-scale constants for three to five components.
For parameter learning, the exact order gives an explicit pointwise maximum-likelihood
exponent and an exact local minimax order on specified shrinking neighbourhoods of a fixed
known mixing measure. The same coefficient viewpoint separates high-order cancellation from
exact parameter symmetries in mixtures of experts. Synchronized replication preserves the
conditional density while keeping the corresponding parameter losses positive in two broad
standard-softmax classes, showing why statistical rate arguments must account for exact
parameter symmetries.}
\newcommand{\notekeywords}{Ho--Nguyen conjecture; Gaussian mixtures; weak identifiability; over-fitted mixtures;
mixtures of experts; softmax identifiability; Gauss--Hermite quadrature; moment matching; pointwise
maximum likelihood; local minimax lower bounds}
\newcommand{\noteMSCprimary}{62F12, 62H30}
\newcommand{\noteMSCsecondary}{62C20, 65D32}

\hypersetup{
  pdftitle={\notetitle},
  pdfauthor={Yuyang Zhang},
  pdfsubject={Gaussian mixture models and mixtures of experts},
  pdfkeywords={\notekeywords}
}

\title{\notetitle}

\author{
  Yuyang Zhang\\[2pt]
  {\small McGill University}\\
  {\small\texttt{yuyang.zhang@mail.mcgill.ca}}
  \and
  Qihuang Zhang\\[2pt]
  {\small McGill University}\\
  {\small\texttt{qihuang.zhang@mcgill.ca}}
}
\date{}

\begin{document}
\maketitle

\begin{abstract}
\noteabstract

\textit{MSC2020:} Primary \noteMSCprimary; secondary \noteMSCsecondary.
\end{abstract}

%
%
%
%
%

\section{Introduction}\label{sec:note-intro}

Learning an over-fitted mixture requires controlling how several fitted components approach
one true component. For Gaussian location--scale mixtures, the Ho--Nguyen coefficient system
encodes the resulting density cancellations. We determine its exact order and use
constructive witnesses, moment rigidity and density-preserving symmetries to connect the
coefficient calculation to parameter identifiability.

\paragraph{Over-fitted mixtures.} When a finite mixture is fitted with more components than the
true number $k_0$, the mixing measure can still be estimated, but at a slower rate that depends on
how the kernel family degenerates. Classical work on strongly identifiable mixtures gives complementary density and mixing-measure
convergence results \citep{chen1995optimal,nguyen2013convergence,ho2016ejs}. For second-order strongly identifiable families, maximum-likelihood estimation achieves the
pointwise $\Wass2$ rate $O_{\PP}((\log n/n)^{1/4})$ under boundedness, regularity,
and entropy conditions
\citep[Theorem~4.3 and Corollary~4.1(b)]{ho2016ejs}. Uniformly
over neighbourhoods of a fixed mixing measure the picture is different: \citet{heinrich2018strong},
who correct a lemma of \citet{chen1995optimal}, show that under their regularity and strong
identifiability conditions the optimal local minimax rate in the transportation distance $\Wass{1}$ is
$n^{-1/(4(k-k_0)+2)}$.
A Bayesian counterpart is
studied by \citet{rousseau2011overfitted}. The location--scale Gaussian family is not strongly
identifiable: its density satisfies the heat equation $\partial^2_\theta f=2\,\partial_\Sigma f$,
so second derivatives in the location are linearly dependent on first derivatives in the scale.
The technical report \citet[Section~4.1, system~(8)]{ho2015arxiv} introduces the integer
$\rbar=\rbar(k-k_0)$, the smallest order at which a polynomial system in $3(k-k_0+1)$ unknowns
has no non-trivial solution (\cref{hn:def-system}), and uses it in a local inverse comparison
between total variation and Wasserstein distance \citep[Theorem~4.1]{ho2015arxiv}.
Appendix~\ref{hn:sec-stat} gives the parameter-space and weight-floor hypotheses of this comparison.
Under their growing-domain hypotheses,
the maximum-likelihood estimator converges at the pointwise rate
$(\log n/n)^{1/(2\rbar)}$ in $\Wass{\rbar}$, while the minimax risk in $\Wass1$ is at least of
order $n^{-1/r}$ for every $r<2\rbar$
\citep[Theorem~1.1(a),(b)]{ho2016aos}. For mixtures of two one-dimensional Gaussians with both mixing weights bounded away from zero,
\citet[Definition~1.1 and Corollaries~1.3 and~1.6]{hardt2015tight} establish tight
sample-complexity bounds for componentwise recovery in their variance-normalised
parameter distance. This differs from uniform $\Wass1$ risk when fitted weights
may vanish.

\paragraph{The conjecture.} \citet[Proposition~4.2]{ho2015arxiv} give
$\rbar(1)=4$, $\rbar(2)=6$, and $\rbar(k-k_0)\ge7$ for $k-k_0\ge3$; the general lower bound also
follows from the Gauss--Hermite witness in \cref{hn:prop-witness}. They conjecture the formula
$\rbar=2(k-k_0+1)$ in the remark following that proposition
\citep[remark after Proposition~4.2]{ho2015arxiv}.
The same system, and close relatives of it, govern parameter estimation in Gaussian mixtures of
experts. For softmax gating, \citet{nguyen2023softmax} established the two-expert value, while
their published three-expert value was accompanied by a defective order-five witness;
Proposition~\ref{hn:prop-witness} supplies a valid order-five witness, while
Corollary~\ref{hn:cor-moe} establishes the full order value. The prediction
$\rbar_{\mathrm{SM}}(m)=2m$ was also recorded by \citet{hai2025dendrogram}. When indexed
by the number of fitted experts, these cases are $\rbar_{\mathrm{SM}}(2)=4$ and
$\rbar_{\mathrm{SM}}(3)=6$;
\citet[Section~3]{nguyen2024topk} state the same numerical values and prediction.
Remark~\ref{hn:rem-literature} explains the different gradings in their equation~(7) and subsequent
derivation. The recent survey by \citet[Section~3.4]{nguyen2026ot} still reports these
known values and lower bound.

\paragraph{Main result.} Write $\Mhn=k-k_0+1$ for the number of fitted atoms that may approach one
true atom. Theorem~\ref{thm:note-hn-body} proves that \HNshortstatement, with the complete statement
and rigidity conclusion given in Section~\ref{sec:note-main}. The examples in
Remark~\ref{hn:rem-hypotheses} give higher-order solutions when the coefficient unknowns
$(a_j,b_j,c_j)$ are allowed jointly to be complex, when weights may be signed, or when zero
weights are allowed. The real rigidity statement equivalently says that a positive mixture of $\Mhn$ univariate
Gaussians cannot match the moments of a single Gaussian through order $2\Mhn$ unless every
component coincides with that Gaussian. The complete proof is given in Appendix~\ref{hn:sec-proof} (see the restatement in Theorem~\ref{hn:thm-main}).

\paragraph{Statistical consequences and structural obstruction.} The coefficient calculation
also gives the exact order $2\Mhn$ for the softmax-gated \citep{nguyen2023softmax},
one-dimensional Gaussian-gated \citep{nguyen2024gaussiangated}, and DeepSeek-type
\citep{nguyen2025deepseek} systems. These orders provide the algebraic input to parameter-rate
arguments. Section~\ref{sec:note-stat} develops the pointwise and local consequences for
location--scale mixtures and the exact synchronized-replication obstructions for two broad
standard-softmax fitted classes. The replications preserve the conditional density while the parameter losses remain positive,
so coercivity fails over these broad fitted classes. Quotient losses and identifiable subclasses
provide natural settings for parameter-rate theory.

\paragraph{Proof architecture.} The lower bound is the positive $\Mhn$-point Gauss--Hermite
moment-matching construction used in related lower bounds \citep{wu2020dmm,diakonikolas2017sq};
for $\Mhn=2$ it recovers the solution in
\citet[Section~4.1, after system~(8)]{ho2015arxiv}. The upper bound is obtained by a
common positive-variance twist followed by the sharp sign-change bound for Gaussian combinations
\citep{kalai2010efficiently,moitra2010settling,karlin1968total}. Section~\ref{sec:note-main}
outlines these three main steps. Appendix~\ref{hn:sec-proof} proves the required
variation-diminishing statement and zero count.

The constructive proof also gives a quantitative experimental prediction.
Section~\ref{sec:note-experiments} follows the Gauss--Hermite path down to
$t=10^{-3}$, then varies the weights and component directions in fixed-size
multistart searches. The prescribed path matches its leading total-variation term,
and the optimized configurations retain the predicted distance powers. This makes
the cancellation mechanism visible and distinguishes its order from its finite-scale constants.

\paragraph{Organisation.} Section~\ref{sec:note-main} states the main theorem and outlines its
proof. Section~\ref{sec:note-stat} develops the statistical consequences and identifiability
obstruction. Section~\ref{sec:note-experiments} tests the cancellation mechanism numerically,
and Section~\ref{sec:note-discussion} develops further directions.
Appendices~\ref{hn:sec-setting}--\ref{hn:sec-remarks} provide the complete proofs,
mixture-of-experts derivations and technical remarks. Appendix~\ref{app:note-experiments}
records the numerical methods and data.

\section{Main theorem and proof architecture}\label{sec:note-main}

We state the Ho--Nguyen coefficient system and outline the three main ingredients of the
proof: Gauss--Hermite quadrature, a common-variance shift, and a sign-change bound.
The complete arguments, including the Gaussian variation-diminishing lemma, appear in Appendix~\ref{hn:sec-proof}.

\begin{definition}[Ho--Nguyen coefficient system]\label{def:note-hn-body}
Let $\Mhn\ge2$ and $r\ge1$. For real unknowns
$(c_j,a_j,b_j)_{j=1}^{\Mhn}$, the system $\hnsys{\Mhn}{r}$ is
\begin{equation}\label{eq:note-hn-body}
\sum_{j=1}^{\Mhn}\ \sum_{\substack{n_1,n_2\ge0\\n_1+2n_2=\alpha}}
\frac{c_j^2a_j^{n_1}b_j^{n_2}}{n_1!\,n_2!}=0,
\qquad \alpha=1,\dots,r.
\end{equation}
A solution is non-trivial when every $c_j$ is non-zero and at least one $a_j$ is non-zero.
The Ho--Nguyen order is the least $r$ for which no such real solution exists.
\end{definition}

\begin{theorem}[Exact Ho--Nguyen order and rigidity]\label{thm:note-hn-body}
\HNmainstatement
\end{theorem}

We defer the detailed proof to Appendix~\ref{hn:sec-proof} (Theorem~\ref{hn:thm-main}) and outline its three steps here. For the existence direction,
Gauss--Hermite quadrature provides positive weights that simultaneously cancel every coefficient
through order $2\Mhn-1$. Let
$x_1<\cdots<x_\Mhn$ be the zeros of the probabilists' Hermite polynomial and let
$\pi_j>0$, $\sum_j\pi_j=1$, be the Gauss--Hermite weights. The quadrature identity
\[
\sum_{j=1}^{\Mhn}\pi_jq(x_j)=\mathbb E q(Z),
\qquad \deg q\le2\Mhn-1,\quad Z\sim\Normal(0,1),
\]
is classical \citep{szego1975orthogonal,golub1969gauss}. With
$c_j=\sqrt{\pi_j}$, $a_j=x_j$, and $b_j=-1/2$, it gives
\[
\hnE_\alpha=0\quad(1\le\alpha\le2\Mhn-1),
\qquad
\hnE_{2\Mhn}=-\frac{\Mhn!}{(2\Mhn)!}.
\]
Thus the witness is non-trivial for $\Mhn\ge2$ and proves sharp existence through order
$2\Mhn-1$. This moment-matching construction is used for minimax lower bounds with unknown
common variance by \citet[Proposition~9]{wu2020dmm}, and for
query-complexity lower bounds in the statistical-query model by
\citet[Lemma~4.3 and Section~4]{diakonikolas2017sq};
at $\Mhn=2$ it is the witness in
\citet[Section~4.1, after system~(8)]{ho2015arxiv}.

For the rigidity direction, the quadratic coefficients may be negative, so the exponentials
need not initially describe Gaussians with positive variances. A common-variance shift removes this obstacle. Put $p_j=c_j^2>0$, $P=\sum_jp_j$, choose
$\bar v>2\max_j|b_j|$, and set $v_j=\bar v+2b_j>0$. Multiplication of the generating function by
$e^{\bar vu^2/2}$ gives the exact equivalence
\[
(c,a,b)\text{ solves }\hnsys{\Mhn}{r}
\quad\Longleftrightarrow\quad
\hnmom_k\!\left(\sum_jp_j\Normal(a_j,v_j)\right)
 =P\,\hnmom_k\!\left(\Normal(0,\bar v)\right),
\quad 0\le k\le r.
\]
Here the shift makes every component variance positive, and Lemma~\ref{hn:lem-twist}
verifies the equivalence by comparing Taylor coefficients of the moment generating functions.

Finally, for rigidity at order $2\Mhn$, subtract the target Gaussian density from the mixture
density. The result is a linear combination of $L=\Mhn+1$ Gaussian
densities. Such a combination has at most $2(L-1)=2\Mhn$ sign changes, whereas a non-zero
continuous function whose moments through order $2\Mhn$ vanish has at least $2\Mhn+1$ sign
changes. Appendix~\ref{hn:sec-proof} proves both statements in Lemmas~\ref{hn:lem-signchange} and
\ref{hn:lem-zerocount}; the latter is derived from the variation-diminishing property of the
Gaussian kernel \citep{karlin1968total}. It also records the alternative zero-count route from
\citet[Proposition~7]{kalai2010efficiently}, as restated in \citet{moitra2010settling}.
Consequently the density difference vanishes. After equal Gaussian components are merged, their
coefficients remain positive, and linear independence forces every $(a_j,v_j)$ to equal
$(0,\bar v)$. Hence $a_j=b_j=0$ for all $j$, establishing rigidity at order $2\Mhn$.

\section{Statistical consequences and the identifiability boundary}\label{sec:note-stat}

For mixtures of experts, the same coefficient method determines the algebraic order.
A statistical rate also requires a coercive comparison between density and parameter loss.
Corollary~\ref{hn:cor-moe} proves order $2\Mhn$ for the softmax-gated system of
\citet{nguyen2023softmax}, the one-dimensional Gaussian-gated system of
\citet{nguyen2024gaussiangated}, and both DeepSeek-type systems of
\citet{nguyen2025deepseek}. The proof restricts each generating function to its Ho--Nguyen
variables for the upper bound and extends the positive Gauss--Hermite witness for the lower bound; the
remaining variables are eliminated by positive sums of squares.

For location--scale mixtures, Fact~\ref{hn:fact-hn-mle} and the exact order give
\[
\Wass{2\Mhn}(\Ghat,G_0)
 =O_{\PP}\!\left(\left(\frac{\log n}{n}\right)^{1/(4\Mhn)}\right).
\]
This is a pointwise maximum-likelihood bound under the hypotheses of
\citet[Theorem~1.1(b)]{ho2016aos}. Their theorem allows fitted weights without a positive lower bound when
$\Mhn\in\{2,3\}$ and requires a fixed positive floor when $\Mhn\ge4$. Proposition~\ref{hn:prop-localminimax} uses a Gauss--Hermite two-point construction inside a
shrinking neighbourhood of a fixed $G_0$ and proves that the minimax risk over that neighbourhood
has exact order $n^{-1/(4\Mhn)}$. The lower-bound
exponent matches the pointwise maximum-likelihood exponent up to its logarithmic factor.

Proposition~\ref{hn:prop-softmax-replication} gives a separate identifiability obstruction. In the
broad standard-softmax class of \citet{nguyen2023softmax}, each of two true experts is
replaced by an unchanged copy and a same-expert copy whose gate slope is shifted by
$v=1/4$. Both the fitted numerator and denominator are multiplied by $1+e^{vx}$, so the
conditional density is unchanged. Both the unrestricted and parameter-preserving common-translation
conventions give the same value in equation~(6) of \citet{nguyen2023softmax}:
\[
D_2^{\mathrm{SM}}=\frac1{1024}>0.
\]
The distinctness condition applies to the true expert triples and
permits these synchronized fitted duplicates. In the dense standard-softmax class of
\citet{nguyen2025deepseek}, the analogous routed replication leaves the shared atom and the
complete conditional density unchanged, but gives
\[
D_2^{\mathrm{DS}}=\frac{257}{128}>0.
\]
This class imposes separation on the true experts and permits coincident fitted
experts. Appendix~C.2 of that paper assumes a fitted reference normalization that is absent from the
definition of the fitted class and Theorem~2.

These examples show that coercivity with respect to the two $D_2$ losses fails on the broad
fitted classes, while the polynomial-system orders retain their algebraic meaning.
For Gaussian-gated mixtures, Remark~\ref{hn:rem-moe-rates} records the candidate parameter
exponents obtained from \citet[Theorems~1 and~2]{nguyen2024gaussiangated} and identifies gaps in
their statistical arguments, including the one-dimensional Type~II setting.

\section{Numerical study of cancellation order}\label{sec:note-experiments}

The constructive proof predicts both an exponent and a leading density term.
We examine these predictions in two stages: first along the prescribed
Gauss--Hermite path, then among configurations found by a multistart search
over weights and component directions. Throughout this section the true
mixing measure is $G_0=\delta_{(0,1)}$, and $K=\Mhn$ denotes the number of
fitted components. The calculations compare population densities directly.

\subsection{The constructive path attains the predicted order}

\begin{figure}[t]
\centering
\includegraphics[width=\textwidth]{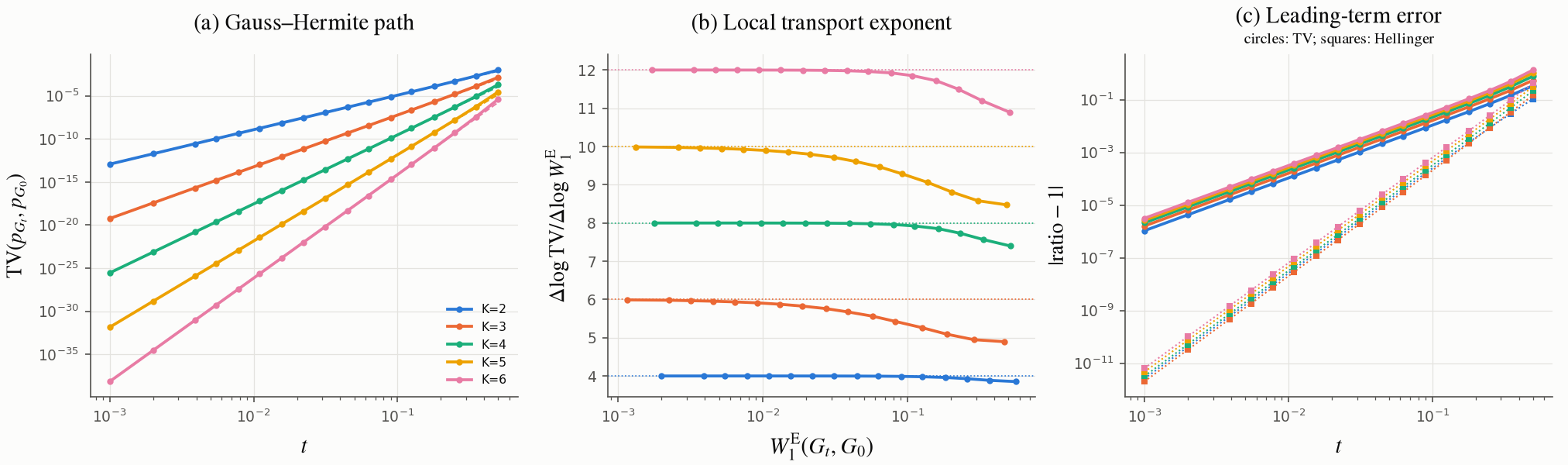}
\caption{The Gauss--Hermite path, evaluated with 110-digit working arithmetic
for $K=2,\ldots,6$. Left: TV against the path parameter $t$; dashed curves
show its theoretical leading term. Middle: local TV slopes against the
Euclidean transport distance $W_1^{\mathrm E}$ on location and variance;
the dotted levels are $2K$. Right: relative deviations from the TV and
Hellinger leading terms, shown by circles and squares, respectively, with
$O(t^2)$ reference behavior for both. The zero middle node for odd $K$
contributes a $t^2$ term to $W_1^{\mathrm E}$ and makes the finite-scale
transport slope converge more slowly.}
\label{fig:note-gh-numerics}
\end{figure}

Let $(x_j,\pi_j)_{j=1}^K$ be the standard-normal Gauss--Hermite rule and set
\[
G_t=\sum_{j=1}^K\pi_j\delta_{(t x_j,\,1-t^2)}.
\]
The weights stay fixed as the components merge. The witness in
\cref{hn:prop-witness} cancels the first $2K-1$ coefficients, and
\cref{hn:lem-gh-tv} identifies the first total-variation term as
\[
\TV(p_{G_t},\hnphi)\sim
\frac{K!}{2(2K)!}\,\EE|\hnHe_{2K}(Z)|\,t^{2K}.
\]
Here $\hnphi$ is the standard-normal density and $\hnHe_r$ is the
probabilists' Hermite polynomial. We evaluated TV and Hellinger distance at 110 decimal digits for
$K=2,\ldots,6$ and 17 fixed scales from $t=0.5$ to $10^{-3}$.
At the smallest scale, the recorded ratio of TV to its predicted leading
term ranges from $1.00000106666759$ to $1.00000316484104$.
All 85 comparisons record $2K$ sign changes of the density difference.
Figure~\ref{fig:note-gh-numerics} shows how the limiting term emerges as
the components approach the truth.

\subsection{Searching over weights and component directions}

We next allow positive weights, locations and variance directions to vary.
For a mixture $G=\sum_jp_j\delta_{(\theta_j,v_j)}$, the homogeneous size is
\[
\|G\|_{\mathrm h}^2
 =\sum_jp_j\sqrt{\theta_j^4+(v_j-1)^2}.
\]
At eight fixed sizes $\|G\|_{\mathrm h}=\varepsilon$, we search for small
$\sqrt{\chisq(p_G\|\hnphi)}$ and independently search for small Hellinger
distance. The variance interval is $[0.1,1.9]$, and the weights are
parameterised by softmax. Starts include the constructive path, random
and perturbed configurations, coefficient-system solutions and
continuation from neighbouring scales. Appendix~\ref{app:note-experiments}
records the full settings and the alternate Euclidean-$W_1$ comparison.

The resulting configurations exhibit the distance exponents in
Table~\ref{tab:note-hn-exponents}. The coefficient objective and the direct
Hellinger objective recover the same orders after high-precision
re-evaluation. For $K=3,4,5$, the searches also find smaller values than the
Gauss--Hermite comparator at the same size. Across these searches, the
observed exponents are stable while the finite-scale constants vary with
the configuration. The vanishing-weight escape of \cref{prop:note-escape}
describes the separate regime of diverging locations.

\begin{table}[t]
\centering
\caption{Observed distance exponents at fixed homogeneous size. Each
reported slope uses the four smallest of eight tested sizes and the saved
high-precision evaluations of the optimized configurations. The last
column is a separate direct-Hellinger search. Its coverage is $K=2,3,4$.}
\label{tab:note-hn-exponents}
\begin{tabular}{@{}rrrr@{}}
\toprule
$K$ & Fixed-weight order $2K$ & $\sqrt{\chisq}$ search & Direct Hellinger search\\
\midrule
2 & 4  & 4.0000 & 4.0001\\
3 & 6  & 6.0000 & 6.0000\\
4 & 8  & 8.0000 & 8.0000\\
5 & 10 & 9.9995 & ---\\
\bottomrule
\end{tabular}
\end{table}

\section{Discussion}\label{sec:note-discussion}

\paragraph{Quadrature, rigidity and symmetry.} The order of the Ho--Nguyen system is $2\Mhn$, so the
location--scale Gaussian mixture with $\Mhn-1$ redundant components behaves, as far as the
exponent is concerned, like a mixture whose first $2\Mhn$ moments are the only obstruction. The
proof identifies the system with a moment-matching problem between a heteroscedastic Gaussian
mixture and one Gaussian, and solves that problem by a sign-change count. Together with Fact~\ref{hn:fact-hn-mle}, this makes the pointwise rate of the
maximum-likelihood estimator explicit. Proposition~\ref{hn:prop-localminimax} separately gives the exact minimax order on
specified shrinking neighbourhoods about a fixed known center.
For standard-softmax and dense standard-softmax mixtures of experts
\citep{nguyen2023softmax,nguyen2025deepseek}, synchronized replication exposes parameter
distinctions invisible to the conditional density. A coercive parameter loss must account for
these exact symmetries, for example through a quotient loss or a restriction of the fitted class.

The numerical study connects the constructive witness to its finite-scale density behaviour.
The prescribed path recovers the leading total-variation term, while the recorded searches
find smaller constants for $K=3,4,5$ and retain the observed cancellation powers. This separates
the exponent settled by rigidity from the quantitative optimization of configurations.

\paragraph{Directions for statistical refinement.} Resolving the algebraic order brings several open statistical questions into sharper focus.
\begin{itemize}[leftmargin=*]
\item \emph{Uniform rates.} Both \cref{hn:cor-mle} and \cref{hn:prop-localminimax} concern a fixed
true mixing measure. The uniform minimax rate over over-fitted location--scale mixtures is open
\citep[Section~3.4]{nguyen2026ot}. The componentwise-recovery formulation of
\citet{hardt2015tight} keeps both mixing weights bounded away from zero; uniform
$\Wass1$ risk at the vanishing-weight boundary is a separate target.
\item \emph{The weight floor.} \Cref{hn:fact-hn-mle} requires a lower bound $c_0>0$ on the fitted
weights when $\Mhn\ge4$. The technical report removes it for $\Mhn\le3$
\citep[Proposition~4.3]{ho2015arxiv}; Corollary~\ref{hn:cor-mle} retains this floor for
$\Mhn\ge4$, and whether it can be removed in general remains open.
\item \emph{Constants.} The Gauss--Hermite arc gives the exponent, but sharp constants in the
local inverse bound
$\TV(p_G,p_{G_0})\gtrsim\Wass{2\Mhn}^{2\Mhn}(G,G_0)$ are not known.
\item \emph{Related systems.} Other polynomial systems of the same flavour remain open. For
two-component Gaussian mixtures with fixed known $\pi\in(0,1/2)$,
\citet{manole2020uniform} prove $\bar r_{\mathrm{asym}}(\pi)\ge6$, verify equality for
$\pi=i/100$ with $1\le i\le49$, and conjecture equality in general. Systems arising for
hierarchical mixtures of experts also remain open. The twist does not apply to these systems
directly.
\item \emph{Separate rates.} The split of rates between locations and scales in
\cref{hn:rem-variance} rests on \citet{manole2022refined} for a penalised estimator; extending
this split rate to the unpenalised maximum-likelihood estimator over $\Okc$ remains open.
\end{itemize}

\section*{Acknowledgements}
%
%

%
%
%
%
Parts of this work, including literature search and proof drafting, were helped by the Danus system. The author takes full responsibility for the
content of the paper.

\clearpage
\appendix
%
%
%
%
%
%
%


\hnsection{Setting and main result}\label{hn:sec-setting}

Let $f(x\mid\theta,\Sigma)$ denote the density of $\Normal(\theta,\Sigma)$ on $\RR^d$. For a
discrete probability measure $G=\sum_{i}p_i\delta_{(\theta_i,\Sigma_i)}$ on
$\RR^d\times\mathbb S^d_{++}$ write $p_G(x)=\sum_ip_if(x\mid\theta_i,\Sigma_i)$. Data
$X_1,\dots,X_n$ are i.i.d.\ from $p_{G_0}$, where the true mixing measure $G_0$ has exactly
$k_0$ atoms, and one fits a mixture with at most $k>k_0$ components. In this
\emph{over-fitted} regime, the technical report
\citet[Section~4.1, system~(8)]{ho2015arxiv} formulates a system of polynomial equations in
$3(k-k_0+1)$ unknowns. Appendix~\ref{hn:sec-stat} gives its statistical consequences
\citep[Theorem~4.1]{ho2015arxiv}. The polynomial system itself has real formal variables and does
not depend on the dimension $d$. In dimension one we write $\hnphi_{\mu,v}$ for the
$\Normal(\mu,v)$ density and $\hnphi=\hnphi_{0,1}$.

\begin{definition}[The Ho--Nguyen system]\label{hn:def-system}
For integers $\Mhn\ge1$ and $r\ge1$, the system $\hnsys{\Mhn}{r}$ in the \emph{real} unknowns
$(c_j,a_j,b_j)_{j=1}^{\Mhn}\in(\RR^3)^{\Mhn}$ is
\begin{equation}\label{hn:eq-system}
\HNsystemdisplay{\Mhn}{r}
\end{equation}
where the inner sum runs over all pairs of non-negative integers $(n_1,n_2)$ with
$n_1+2n_2=\alpha$. A solution is \emph{non-trivial} if $c_j\neq0$ for all $j$ and $a_j\neq0$
for some $j$. For $k>k_0\ge1$, the \emph{Ho--Nguyen order} $\rbar(k-k_0)$ is the least $r\ge1$
for which $\hnsys{k-k_0+1}{r}$ has no non-trivial real solution.
\end{definition}

This is system~(8) of \citet[Section~4.1]{ho2015arxiv}; see \cref{hn:rem-numbering}. Every solution of
$\hnsys{\Mhn}{r+1}$ solves $\hnsys{\Mhn}{r}$, so solvability is downward closed in $r$ and
$\rbar(k-k_0)=1+\max\{r:\hnsys{k-k_0+1}{r}\text{ has a non-trivial real solution}\}$. Without the
condition $c_j\neq0$ the order would not exist (see the remark following system~(8)
in that report).

\paragraph{Generating function.} Collecting the equations in one power series lets us
track their cancellation under a common variance shift. For $a,b\in\RR$ and $\alpha\ge0$ put
$\hne_\alpha(a,b)=\sum_{n_1+2n_2=\alpha}a^{n_1}b^{n_2}/(n_1!\,n_2!)$. Multiplying the
exponential series of $au$ and $bu^2$ gives
\begin{equation}\label{hn:eq-genfun}
\exp(au+bu^2)=\sum_{\alpha\ge0}\hne_\alpha(a,b)\,u^\alpha ,\qquad
\hne_\alpha(\tau a,\tau^2b)=\tau^\alpha\hne_\alpha(a,b)\quad(\tau\in\RR).
\end{equation}
Hence the $\alpha$-th equation of \eqref{hn:eq-system} reads $\hnE_\alpha=0$, where
\begin{equation}\label{hn:eq-Ealpha}
\Phi(u):=\sum_{j=1}^{\Mhn}c_j^2\exp(a_ju+b_ju^2)=\sum_{\alpha\ge0}\hnE_\alpha u^\alpha ,\qquad
\hnE_\alpha=\hnE_\alpha(c,a,b):=\sum_{j=1}^{\Mhn}c_j^2\,\hne_\alpha(a_j,b_j).
\end{equation}
Thus $(c,a,b)$ solves $\hnsys{\Mhn}{r}$ if and only if
$\Phi(u)\equiv\sum_jc_j^2\pmod{u^{r+1}}$ in $\RR[[u]]$. By the homogeneity in
\eqref{hn:eq-genfun}, the solution set of $\hnsys{\Mhn}{r}$ is invariant under
$(c,a,b)\mapsto(\tau'c,\tau a,\tau^2b)$ for $\tau'\neq0$ and $\tau\in\RR$.

\begin{remark}[Real unknowns]\label{hn:rem-real}
\citet{ho2015arxiv} specify a ``non-trivial real solution'' in their introductory
definition, while the detailed definition following system~(8) in Section~4.1 does not
repeat ``real''. The unknowns arise as limits of normalised differences of real
parameters in the proof of their density bound. Their proof of Proposition~4.2 shows
that a normalised system for $\rbar(1)=4$ has no \emph{real} solution, using the
Gr\"obner basis element $x^6+2x^4+2x^2+1>0$.
\citet{manole2022refined} state the definition with $(a_j,b_j,c_j)\subseteq\RR$ explicitly. Over
$\CC$ the conclusion of \cref{hn:thm-main} is false (\cref{hn:rem-hypotheses}).
\end{remark}

\begin{theorem}[Ho--Nguyen order; restatement]\label{hn:thm-main}
For $\Mhn\ge2$ and $r\ge1$, the real system of Definition~\ref{hn:def-system} has the exact
solvability threshold and rigidity property stated in Theorem~\ref{thm:note-hn-body}.
\end{theorem}

The proof in Appendix~\ref{hn:sec-proof} proceeds through the Gauss--Hermite witness
(\cref{hn:prop-witness}), the common-variance shift (\cref{hn:lem-twist}), and the Gaussian
sign-change bound (\cref{hn:lem-zerocount,hn:prop-moment}).

\begin{remark}[Stronger form]\label{hn:rem-strong}
For every $\Mhn\ge1$, \emph{every} real solution of $\hnsys{\Mhn}{2\Mhn}$ satisfies
$c_ja_j=c_jb_j=0$ for all $j$: atoms of non-zero weight sit at the origin. Indeed, if
$\Mhn'\ge1$ atoms have $c_j\neq0$, they solve $\hnsys{\Mhn'}{2\Mhn}$, hence
$\hnsys{\Mhn'}{2\Mhn'}$, and the rigidity part of \cref{hn:thm-main}, whose proof below does not use
$\Mhn\ge2$, applies. The restriction $\Mhn\ge2$ is needed only for the witness, since the one-point
rule has its node at $0$.
\end{remark}

\hnsection{Proof of the main theorem}\label{hn:sec-proof}

Write $\hnmom_k(\mu)=\int x^k\,\mu(\dd x)$ for the moments of a finite measure $\mu$ on $\RR$.
For $Y\sim\Normal(a,v)$ the moment generating function $\EE e^{uY}=\exp(au+vu^2/2)$ is entire,
with Taylor coefficients $\hnmom_k(\Normal(a,v))/k!$.

\hnsubsection{The twist}\label{hn:subsec-twist}

A common positive-variance shift converts the polynomial system into a Gaussian
moment-matching problem.

\begin{lemma}[Twist: the Ho--Nguyen system is moment matching]\label{hn:lem-twist}
Let $(c,a,b)\in(\RR^3)^{\Mhn}$ with all $c_j\neq0$, put $p_j=c_j^2>0$ and $P=\sum_jp_j$, and let
$\bar v>0$ satisfy $v_j:=\bar v+2b_j>0$ for all $j$ (for instance $\bar v>2\max_j|b_j|$). Let
$\mu=\sum_jp_j\Normal(a_j,v_j)$. Then $(c,a,b)$ solves $\hnsys{\Mhn}{r}$ if and only if
$\hnmom_k(\mu)=P\,\hnmom_k(\Normal(0,\bar v))$ for $k=0,1,\dots,r$.
\end{lemma}

\begin{proof}
By \eqref{hn:eq-Ealpha}, $(c,a,b)$ solves $\hnsys{\Mhn}{r}$ if and only if
$\Phi(u)-P\in u^{r+1}\RR[[u]]$. Since $e^{\bar vu^2/2}$ is a unit of $\RR[[u]]$, this holds if and only
if $e^{\bar vu^2/2}\Phi(u)-Pe^{\bar vu^2/2}\in u^{r+1}\RR[[u]]$. Now
\[
e^{\bar vu^2/2}\Phi(u)-Pe^{\bar vu^2/2}
=\sum_jp_j\exp\Big(a_ju+\tfrac{v_j}{2}u^2\Big)-P\exp\Big(\tfrac{\bar v}2u^2\Big)
=\sum_{k\ge0}\frac{u^k}{k!}\Big(\hnmom_k(\mu)-P\,\hnmom_k(\Normal(0,\bar v))\Big),
\]
because the two exponentials are the moment generating functions of $\Normal(a_j,v_j)$ and
$\Normal(0,\bar v)$. The coefficient of $u^0$ vanishes ($\mu$ has mass $P$), so the condition is
exactly $\hnmom_k(\mu)=P\,\hnmom_k(\Normal(0,\bar v))$ for $0\le k\le r$.
\end{proof}

For $r\ge2$, every non-trivial real solution of $\hnsys{\Mhn}{r}$
has some $b_j<0$ (a variance-decreasing direction): if all $b_j\ge0$, the imposed
second equation $\hnE_2=\sum_jp_j(a_j^2/2+b_j)=0$, with $p_j=c_j^2>0$,
forces $a_j=b_j=0$ for every $j$, contradicting non-triviality. The twist absorbs
such directions into the common variance $\bar v$.

\hnsubsection{The Gauss--Hermite witness}\label{hn:subsec-witness}

Let $Z\sim\Normal(0,1)$ and let $\hnHe_n(x)=(-1)^ne^{x^2/2}\frac{\dd^n}{\dd x^n}e^{-x^2/2}$ be
the probabilists' Hermite polynomials.

For existence, the nodes and positive weights of classical Gauss--Hermite quadrature give
a non-trivial solution through order $2\Mhn-1$.

\begin{lemma}[Gauss--Hermite rule]\label{hn:lem-gh}
Let $\Mhn\ge1$.
\begin{enumerate}[label=(\roman*)]
\item $\hnHe_\Mhn$ is monic of degree $\Mhn$, $\hnHe_\Mhn(-x)=(-1)^\Mhn\hnHe_\Mhn(x)$, and
$\EE[\hnHe_\Mhn(Z)q(Z)]=\EE[q^{(\Mhn)}(Z)]$ for every polynomial $q$. In particular
$\EE[\hnHe_\Mhn(Z)q(Z)]=0$ if $\deg q<\Mhn$, and $\EE\,\hnHe_\Mhn(Z)^2=\Mhn!$.
\item $\hnHe_\Mhn$ has $\Mhn$ distinct real zeros $x_1<\dots<x_\Mhn$, placed symmetrically about $0$.
\item There are weights $\pi_1,\dots,\pi_\Mhn>0$ with $\sum_j\pi_j=1$ such that
$\sum_j\pi_jq(x_j)=\EE\,q(Z)$ for every polynomial $q$ of degree at most $2\Mhn-1$.
\end{enumerate}
\end{lemma}

\begin{proof}
(i) With $\hnphi$ the standard normal density, $\hnphi^{(n)}=(-1)^n\hnHe_n\hnphi$; differentiating
once more gives $\hnHe_{n+1}(x)=x\hnHe_n(x)-\hnHe_n'(x)$ with $\hnHe_0=1$, so by induction
$\hnHe_n$ is monic of degree $n$ and has the parity of $n$. Integrating by parts $\Mhn$ times
(the boundary terms are polynomials times derivatives of $\hnphi$ and vanish at $\pm\infty$),
$\EE[\hnHe_\Mhn(Z)q(Z)]=(-1)^\Mhn\int q\,\hnphi^{(\Mhn)}=\int q^{(\Mhn)}\hnphi$. For
$q=\hnHe_\Mhn$ this equals $\Mhn!$.
(ii) Let $y_1<\dots<y_m$ be the real zeros of odd multiplicity of $\hnHe_\Mhn$ and
$Q(x)=\prod_i(x-y_i)$. Then $\hnHe_\Mhn Q$ has constant sign and is not identically zero, so
$\EE[\hnHe_\Mhn(Z)Q(Z)]\neq0$, and (i) forces $m=\deg Q\ge\Mhn$. Hence all $\Mhn$ zeros are real
and simple; the symmetry follows from the parity.
(iii) Let $\ell_j$ be the Lagrange polynomials of degree $\Mhn-1$ at the nodes $x_j$, and
$\pi_j=\EE\,\ell_j(Z)$. If $\deg q\le2\Mhn-1$, division gives $q=\hnHe_\Mhn u+w$ with
$\deg u,\deg w\le\Mhn-1$; by (i) and the exactness of interpolation,
$\EE q(Z)=\EE w(Z)=\sum_j\pi_jw(x_j)=\sum_j\pi_jq(x_j)$. Taking $q=\ell_j^2$ gives
$\pi_j=\EE\,\ell_j(Z)^2>0$, and $q=1$ gives $\sum_j\pi_j=1$.
\end{proof}

These facts are classical \citep{szego1975orthogonal}; the nodes and weights are computed
stably by the method of \citet{golub1969gauss}.

\begin{proposition}[Gauss--Hermite witness]\label{hn:prop-witness}
Let $\Mhn\ge1$ and let $(x_j,\pi_j)_{j\le\Mhn}$ be the rule of \cref{hn:lem-gh}. Put
$c_j=\sqrt{\pi_j}$, $a_j=x_j$ and $b_j=-\frac12$. Then $\hnE_\alpha(c,a,b)=0$ for
$1\le\alpha\le2\Mhn-1$ and
\[
\hnE_{2\Mhn}(c,a,b)=-\frac{\Mhn!}{(2\Mhn)!},\qquad\text{equivalently}\qquad
\sum_j\pi_jx_j^{2\Mhn}-\EE Z^{2\Mhn}=-\Mhn!.
\]
For $\Mhn\ge2$ the solution is non-trivial, so $\hnsys{\Mhn}{r}$ has a non-trivial real solution
for every $r\le2\Mhn-1$.
\end{proposition}

\begin{proof}
Let $D_n=\sum_j\pi_jx_j^n-\EE Z^n$; by \cref{hn:lem-gh}(iii), $D_n=0$ for $n\le2\Mhn-1$. Since
$\sum_j\pi_je^{x_ju}$ and $\EE e^{uZ}=e^{u^2/2}$ are entire with Taylor coefficients
$\sum_j\pi_jx_j^n/n!$ and $\EE Z^n/n!$,
\[
\Phi(u)=e^{-u^2/2}\sum_j\pi_je^{x_ju}
=e^{-u^2/2}\Big(e^{u^2/2}+\sum_{n\ge2\Mhn}\frac{D_n}{n!}u^n\Big)
=1+\frac{D_{2\Mhn}}{(2\Mhn)!}u^{2\Mhn}+O(u^{2\Mhn+1}).
\]
Hence $\hnE_0=1$, $\hnE_\alpha=0$ for $1\le\alpha\le2\Mhn-1$, and
$\hnE_{2\Mhn}=D_{2\Mhn}/(2\Mhn)!$. Write $x^{2\Mhn}=\hnHe_\Mhn(x)^2+q(x)$ with $\deg q\le2\Mhn-1$
($\hnHe_\Mhn$ is monic). As $\hnHe_\Mhn(x_j)=0$, \cref{hn:lem-gh} gives
$\sum_j\pi_jx_j^{2\Mhn}=\EE q(Z)$ and $\EE Z^{2\Mhn}=\Mhn!+\EE q(Z)$, so $D_{2\Mhn}=-\Mhn!$. All
$c_j\neq0$; for $\Mhn\ge2$ the nodes are $\Mhn$ distinct reals, so some $a_j\neq0$.
\end{proof}

For $\Mhn=2$ the witness is $c_1=c_2$, $a=(1,-1)$, $b=(-\frac12,-\frac12)$, the solution of
order $3$ exhibited in \citet[Section~4.1, after system~(8)]{ho2015arxiv}.

\hnsubsection{Sign changes and a zero count}\label{hn:subsec-zero}

For $f:\RR\to\RR$ let $\hnsc(f)$ be the supremum, over all finite increasing sequences
$x_1<\dots<x_m$, of the number of sign changes of $(f(x_1),\dots,f(x_m))$, zero entries being
discarded.

Rigidity follows by comparing the sign changes forced by vanishing moments with the upper
bound for a linear combination of Gaussian densities.

\begin{lemma}[Vanishing moments force sign changes]\label{hn:lem-signchange}
Let $r$ be a non-negative integer and let $f:\RR\to\RR$ be continuous and not identically zero,
with $\int|x|^r|f(x)|\,\dd x<\infty$ and $\int x^kf(x)\,\dd x=0$ for
$k=0,1,\dots,r$. Then $\hnsc(f)\ge r+1$.
\end{lemma}

\begin{proof}
Suppose $S:=\hnsc(f)\le r$. Since $f\not\equiv0$ there are $x_0<\dots<x_S$ and
$\varepsilon\in\{\pm1\}$ with $(-1)^i\varepsilon f(x_i)>0$ for all $i$. For $1\le i\le S$ let
$t_i=\sup\{x\in[x_{i-1},x_i]:(-1)^{i-1}\varepsilon f(x)>0\}$; by continuity $f(t_i)=0$ and
$x_{i-1}<t_i<x_i$. Put $t_0=-\infty$, $t_{S+1}=+\infty$ and $\sigma(x)=(-1)^j\varepsilon$ on
$(t_j,t_{j+1})$. We claim $\sigma f\ge0$ off $\{t_1,\dots,t_S\}$. Let $y\in(t_j,t_{j+1})$ with
$\sigma(y)f(y)<0$. If $y<x_0$ or $y>x_S$, adjoining $y$ to $(x_i)$ produces $S+1$ sign changes. If
$t_j<y<x_j$ with $j\ge1$, then $(-1)^{j-1}\varepsilon f(y)\le0$ by the definition of $t_j$, which
contradicts $\sigma(y)f(y)<0$. If $x_j<y<t_{j+1}$, the definition of $t_{j+1}$ yields
$z\in(y,x_{j+1}]$ with $(-1)^j\varepsilon f(z)>0$, and $x_0,\dots,x_j,y,z,x_{j+1},\dots,x_S$ has
$S+2$ sign changes. Each case contradicts $\hnsc(f)=S$. Now
$P(x)=\varepsilon(-1)^S\prod_{i=1}^S(x-t_i)$ has the sign of $\sigma$ on each $(t_j,t_{j+1})$, so
$Pf\ge0$ off a finite set, while $\int Pf=0$ because $\deg P=S\le r$. Hence $Pf=0$ almost
everywhere, and $f\equiv0$ by continuity, a contradiction.
\end{proof}

\begin{lemma}[Linear independence]\label{hn:lem-linind}
Let $(\mu_i,v_i)\in\RR\times(0,\infty)$, $i=1,\dots,L$, be distinct and $\beta_i\in\RR$. If
$\sum_i\beta_i\hnphi_{\mu_i,v_i}\equiv0$ on $\RR$, then $\beta_1=\dots=\beta_L=0$.
\end{lemma}

\begin{proof}
Integrating against $e^{\xi x}$ gives $\sum_i\beta_i\exp(\mu_i\xi+v_i\xi^2/2)=0$ for all
$\xi\in\RR$. If some $\beta_i\neq0$, let $i^*$ maximise $(v_i,\mu_i)$ lexicographically among
those indices. Dividing by $\exp(\mu_{i^*}\xi+v_{i^*}\xi^2/2)$ and letting $\xi\to+\infty$, every
other term tends to $0$, so $\beta_{i^*}=0$, a contradiction.
\end{proof}

The upper bound needs a sharp count of the sign changes of Gaussian combinations. It rests on the
variation-diminishing property of the Gaussian kernel, a classical consequence of its total
positivity \citep[Chapter~5]{karlin1968total}. To keep the argument self-contained we prove the form
that we need, by approximating the Gaussian kernel with convolutions of one-sided exponential kernels.
For a finite signed Borel measure $\nu$ on $\RR$ let $\hnsc(\nu)$ be the supremum of the integers
$k\ge0$ for which there are Borel sets $B_0,\dots,B_k$, each lying strictly to the left of the next
(every point of $B_{j-1}$ is smaller than every point of $B_j$), and a sign $\varepsilon\in\{\pm1\}$
with $\varepsilon(-1)^j\nu(B_j)>0$ for all $j$; put $\hnsc(0)=0$.

\begin{lemma}[Gaussian smoothing does not increase sign changes]\label{hn:lem-vd}
Let $\nu$ be a finite signed Borel measure on $\RR$.
\begin{enumerate}[label=(\roman*)]
\item If $\nu=F\sigma$ for a positive Borel measure $\sigma$ and a $\sigma$-integrable Borel function $F$,
then $\hnsc(\nu)\le\hnsc(F)$.
\item For every $c>0$ the function $g(x)=\int\hnphi_{0,c}(x-y)\,\nu(\dd y)$ satisfies
$\hnsc(g)\le\hnsc(\nu)$.
\end{enumerate}
\end{lemma}

\begin{proof}
(i) If $\varepsilon(-1)^j\nu(B_j)=\int_{B_j}\varepsilon(-1)^jF\,\dd\sigma>0$, there is a point $y_j\in B_j$
with $\varepsilon(-1)^jF(y_j)>0$. The points satisfy $y_0<\dots<y_k$, and $F$ changes sign $k$ times
along them, so $\hnsc(F)\ge k$.

(ii) We use four elementary steps.

\emph{(a) Positive densities.} Let $\varrho>0$ be a Borel function and $\nu'=\varrho\,\nu$, and let
$B_0,\dots,B_k$ be strictly ordered Borel sets on which $\nu'$ is finite, with
$\varepsilon(-1)^j\nu'(B_j)>0$. Let $\RR=P\cup N$ be a Hahn decomposition for $\nu$. If $\nu'(B_j)>0$,
then $\nu'(B_j\cap P)>0$ because $\nu'(B_j\cap N)\le0$, and hence $\nu(B_j\cap P)>0$; if
$\nu'(B_j)<0$, then likewise $\nu(B_j\cap N)<0$. The sets $B_j\cap P$ or $B_j\cap N$ are again
strictly ordered, so $\nu$ has the same $k$ sign changes.

\emph{(b) One-sided exponential kernels.} For $a>0$ let $e_a^+(x)=a^{-1}e^{-x/a}\one\{x>0\}$ and
$e_a^-(x)=e_a^+(-x)$. We claim $\hnsc(e_a^\pm*\nu)\le\hnsc(\nu)$, where
$(e_a^+*\nu)(x)=a^{-1}\int_{(-\infty,x)}e^{-(x-y)/a}\nu(\dd y)$. Put $\nu'=a^{-1}e^{y/a}\nu$, which is finite
on every set bounded above, and $G(x)=\nu'((-\infty,x))$, so that $(e_a^+*\nu)(x)=e^{-x/a}G(x)$ has the
sign of $G(x)$. Suppose $\varepsilon(-1)^jG(x_j)>0$ for $x_0<\dots<x_k$. Put $B_0=(-\infty,x_0)$ and
$B_j=[x_{j-1},x_j)$ for $1\le j\le k$. Then $\nu'(B_0)=G(x_0)$ and $\nu'(B_j)=G(x_j)-G(x_{j-1})$, a
difference of a number of sign $\varepsilon(-1)^j$ and a number of the opposite sign, so
$\varepsilon(-1)^j\nu'(B_j)>0$ for all $j$. By (a), $\hnsc(\nu)\ge k$. The case of $e_a^-$ is the
mirror image, with $\nu'=a^{-1}e^{-y/a}\nu$, $H(x)=\nu'((x,\infty))$ and the sets $(x_j,x_{j+1}]$,
$(x_k,\infty)$.

\emph{(c) Laplace kernels.} The Laplace density $\ell_a(x)=(2a)^{-1}e^{-|x|/a}$ equals $e_a^+*e_a^-$, and
its characteristic function is $(1+a^2\xi^2)^{-1}$. For a finite signed measure $\nu$, the function
$g_1=e_a^-*\nu$ is integrable, and $\ell_a*\nu=e_a^+*(g_1\,\dd x)$ by Fubini's theorem. By (b), (i) with
$\sigma$ Lebesgue measure, and (b) again,
$\hnsc(\ell_a*\nu)\le\hnsc(g_1\,\dd x)\le\hnsc(g_1)\le\hnsc(\nu)$. By induction,
$\hnsc(\ell_a^{*n}*\nu)\le\hnsc(\nu)$ for every $n\ge1$.

\emph{(d) Approximation.} Let $K_n=\ell_{a_n}^{*n}$ with $a_n=(c/(2n))^{1/2}$. Its characteristic function
$(1+c\xi^2/(2n))^{-n}$ converges to $e^{-c\xi^2/2}$ and is bounded by $(1+c\xi^2/2)^{-1}$, which is
integrable, because $(1+x/n)^n\ge1+x$ for $x\ge0$. By Fourier inversion and dominated convergence,
$\sup_x|K_n(x)-\hnphi_{0,c}(x)|\le(2\pi)^{-1}\int|(1+c\xi^2/(2n))^{-n}-e^{-c\xi^2/2}|\,\dd\xi\to0$, and
therefore $|(K_n*\nu)(x)-g(x)|\le\sup|K_n-\hnphi_{0,c}|\,|\nu|(\RR)\to0$ for every $x$. If $g$ has
alternating non-zero signs at $x_0<\dots<x_k$, so has $K_n*\nu$ for all large $n$. Hence
$\hnsc(g)\le\sup_n\hnsc(K_n*\nu)\le\hnsc(\nu)$ by (c).
\end{proof}

\begin{lemma}[Zero count]\label{hn:lem-zerocount}
Let $L\ge1$, $\beta_i\in\RR$ and $(\mu_i,v_i)\in\RR\times(0,\infty)$ for $i=1,\dots,L$, and put
$f=\sum_{i=1}^L\beta_i\hnphi_{\mu_i,v_i}$. Then $\hnsc(f)\le2(L-1)$. No distinctness of the pairs
and no condition on the signs of the $\beta_i$ is required.
\end{lemma}

\begin{proof}
Induction on $L$. Let $v_{\min}=\min_iv_i$ and $I_0=\{i:v_i=v_{\min}\}\neq\emptyset$. For
$i\notin I_0$, $\hnphi_{\mu_i,v_i}=\hnphi_{0,v_{\min}}*\hnphi_{\mu_i,v_i-v_{\min}}$, and for
$i\in I_0$, $\hnphi_{\mu_i,v_{\min}}=\hnphi_{0,v_{\min}}*\delta_{\mu_i}$. Hence
\[
f(x)=\int\hnphi_{0,v_{\min}}(x-y)\,\nu(\dd y),\qquad \nu=F\sigma,\qquad
\sigma=\mathrm{Leb}+\sum_{\mu\in A}\delta_{\mu},\qquad A=\{\mu_i:i\in I_0\},
\]
where $F=h:=\sum_{i\notin I_0}\beta_i\hnphi_{\mu_i,v_i-v_{\min}}$ off $A$ and
$F(\mu)=\sum_{i\in I_0:\,\mu_i=\mu}\beta_i$ for $\mu\in A$ (redefining $F$ on the Lebesgue-null set
$A$ does not affect the Lebesgue integral). The measure $\nu$ is finite because $h$ is integrable, so
\cref{hn:lem-vd} gives $\hnsc(f)\le\hnsc(\nu)\le\hnsc(F)$. Changing one entry of a finite sequence
increases its number of sign changes by at most $2$ (deleting an entry cannot increase it, and
inserting one adds at most $2$). Since $F$ and $h$ differ at most at the $|A|\le|I_0|$ points of $A$,
$\hnsc(F)\le\hnsc(h)+2|I_0|$. If $I_0=\{1,\dots,L\}$, then $h\equiv0$ and
$\hnsc(F)\le|A|-1\le2(L-1)$; this covers $L=1$. Otherwise $h$ is a combination of
$L-|I_0|\ge1$ Gaussian densities, the induction hypothesis gives $\hnsc(h)\le2(L-|I_0|-1)$, and
$\hnsc(f)\le2(L-|I_0|-1)+2|I_0|=2(L-1)$.
\end{proof}

An elementary proof of the distinct-zero bound removes the smallest variance by
deconvolution, leaving a Gaussian combination and one point mass.

\begin{lemma}[Exponential polynomials and Gaussian smoothing with one atom]
\label{hn:lem-exp-atom-zeros}
\begin{enumerate}[label=(\roman*)]
\item Let $n\ge1$ be an integer and let $\lambda_1<\dots<\lambda_n$ be real. The functions
$e^{\lambda_j y}$ are linearly independent, and every nonzero real combination
$Q(y)=\sum_{j=1}^nc_je^{\lambda_j y}$ has at most $n-1$ real zeros counted with multiplicity.
\item Let $v>0$, let $h$ be a nonzero finite real linear combination of Gaussian densities with
finitely many real zeros, and let $y_1<\dots<y_S$ be exactly the zeros at which $h$ changes sign.
For $A,\mu\in\RR$, the function
\[
g=\hnphi_{0,v}*(h(y)\,\dd y+A\delta_\mu)
\]
has at most $S+2$ distinct real zeros. Here $\delta_\mu$ is the unit point mass at $\mu$.
\end{enumerate}
\end{lemma}

\begin{proof}
(i) We prove linear independence and the zero bound together by induction on $n$. Both claims
hold for $n=1$. For the induction step, set
\[
R(y)=e^{-\lambda_1y}Q(y),\qquad
R'(y)=\sum_{j=2}^nc_j(\lambda_j-\lambda_1)e^{(\lambda_j-\lambda_1)y}.
\]
If $Q$ vanishes identically, so does $R'$. Linear independence for $n-1$ exponentials gives
$c_j=0$ for $j\ge2$, and then $c_1=0$. This proves linear independence for $n$.

Now let $Q$ be nonzero. Multiplication by $e^{-\lambda_1y}$ preserves zeros and their
multiplicities. If $R'\equiv0$, then $R$ is a nonzero constant and has no zeros. Otherwise,
$R'$ is a nonzero combination of at most $n-1$ exponentials with distinct real exponents, so
the induction hypothesis bounds its zeros, counted with multiplicity, by $n-2$. A finite
selection of zeros of $R$ with total multiplicity $N$ gives at least $N-1$ zeros of $R'$:
a selected zero of multiplicity $r$ contributes $r-1$ at that point, and Rolle's theorem
gives another zero between consecutive selected points. Thus $N-1\le n-2$, or $N\le n-1$.
This finite-selection argument also excludes an infinite zero set.

(ii) Suppose, to obtain a contradiction, that $g$ has distinct zeros
$x_1<\dots<x_{S+3}$. Put $z_j=x_j/v$. Expanding the square in the Gaussian kernel gives
\[
g(x)=\frac{e^{-x^2/(2v)}}{\sqrt{2\pi v}}L(x/v),\qquad
L(z)=\int_{\RR}e^{zy-y^2/(2v)}h(y)\,\dd y
     +Ae^{z\mu-\mu^2/(2v)}.
\]
Every integral is absolutely convergent: after expanding $h$, each summand has a negative
quadratic coefficient in $y$. In particular $L(z_j)=0$ for every $j$.

Choose real coefficients $c_1,\dots,c_{S+3}$, not all zero, so that
\[
Q(y)=\sum_{j=1}^{S+3}c_je^{z_jy}
\]
satisfies the following homogeneous conditions. If $\mu$ is distinct from every $y_i$, impose
$Q(y_i)=0$ for $1\le i\le S$ and $Q(\mu)=Q'(\mu)=0$. If $\mu=y_r$ for some $r$, impose
$Q(y_i)=0$ for $i\ne r$ and $Q(\mu)=Q'(\mu)=Q''(\mu)=0$. In either case there are $S+2$
conditions on $S+3$ unknowns, so a nonzero coefficient vector exists. The exponents $z_j$ are
distinct, so part (i) makes $Q$ nonzero and bounds its real zeros, counted with multiplicity,
by $S+2$. The imposed zeros already account for that total. They therefore have exactly the
imposed multiplicities, and $Q$ has no further real zeros.

It follows that $Q$ changes sign exactly at the points $y_i$: the separate zero at $\mu$ has
multiplicity two, or the zero at $\mu=y_r$ has multiplicity three. The function $h$ has the same
sign-change pattern. After replacing $Q$ by $-Q$ if needed, we have $Qh\ge0$ on $\RR$, with
strict inequality on a nonempty open interval. Both functions are nonzero and analytic, and
their zero sets here are finite. The product $Q(y)e^{-y^2/(2v)}h(y)$ is absolutely integrable
by the same negative-quadratic estimate. Since $Q(\mu)=0$, finite summation and integration give
\[
\begin{aligned}
0=\sum_{j=1}^{S+3}c_jL(z_j)
 &=\int_{\RR}Q(y)e^{-y^2/(2v)}h(y)\,\dd y
   +AQ(\mu)e^{-\mu^2/(2v)}\\
 &=\int_{\RR}Q(y)e^{-y^2/(2v)}h(y)\,\dd y>0,
\end{aligned}
\]
a contradiction. The argument includes $S=0$ and proves the second claim.
\end{proof}

\begin{fact}[{\citealp[Proposition~7]{kalai2010efficiently}, restated in
\citealp[Proposition~19]{moitra2010settling}}]\label{hn:fact-kmv}
Let $L\ge1$, $\beta_i,\mu_i\in\RR$, and $v_i>0$ for $1\le i\le L$. Suppose that the $v_i$ are
pairwise distinct and that the $\beta_i$ are not all zero. Then
$f=\sum_{i=1}^L\beta_i\hnphi_{\mu_i,v_i}$ has at most $2(L-1)$ distinct real zeros.
\end{fact}

\begin{proof}
Drop terms whose coefficients vanish. If $\ell\le L$ terms remain, a bound of $2(\ell-1)$
suffices, so assume that every $\beta_i$ is nonzero. We induct on the number of remaining terms.
A single Gaussian density has no real zero. For $L\ge2$, relabel the terms so that $v_L=v$ is
the unique smallest variance. The Gaussian convolution identity gives
\[
f=\hnphi_{0,v}*(h(y)\,\dd y+\beta_L\delta_{\mu_L}),\qquad
h(y)=\sum_{i=1}^{L-1}\beta_i\hnphi_{\mu_i,v_i-v}(y).
\]
The variances $v_i-v$ are positive and pairwise distinct. To see directly that $h$ is nonzero,
choose the index $r$ with largest $v_r-v$. For $i\ne r$, the ratio
\[
\frac{\hnphi_{\mu_i,v_i-v}(y)}{\hnphi_{\mu_r,v_r-v}(y)}
\]
is a positive constant times the exponential of a quadratic polynomial whose quadratic
coefficient is
\[
\frac{1}{2(v_r-v)}-\frac{1}{2(v_i-v)}<0.
\]
Thus every such ratio tends to zero as $y\to+\infty$, and
$h(y)/\hnphi_{\mu_r,v_r-v}(y)\to\beta_r\ne0$.

By induction $h$ has at most $2(L-2)$ distinct real zeros. Its zero set is therefore finite,
and the number $S$ of its sign-changing zeros is at most $2(L-2)$. Part (ii) of
\cref{hn:lem-exp-atom-zeros} bounds the distinct zeros of $f$ by
$S+2\le2(L-2)+2=2(L-1)$, completing the induction.
\end{proof}

\begin{remark}[A second route to the zero count]\label{hn:rem-kmv}
The main proof uses Lemma~\ref{hn:lem-vd}; the exponential-polynomial argument in
\cref{hn:fact-kmv} gives another route to \cref{hn:lem-zerocount}. Merge
the terms of $f$ with equal pairs and drop zero coefficients, so that
$f=\sum_{\ell=1}^{L'}\beta_\ell'\hnphi_{\mu_\ell',v_\ell'}$ with $L'\le L$, distinct pairs and all
$\beta_\ell'\neq0$ (if $L'=0$ there is nothing to prove). Let $x_0<\dots<x_S$ with
$f(x_{i-1})f(x_i)<0$ for $1\le i\le S$, and $\delta=\min_i|f(x_i)|>0$. Choose $\eta>0$ so small that
the numbers $v_\ell'+\ell\eta$, $\ell=1,\dots,L'$, are pairwise distinct (this excludes finitely many
values of $\eta$) and $|f_\eta(x_i)-f(x_i)|<\delta$ for all $i$, where
$f_\eta=\sum_\ell\beta_\ell'\hnphi_{\mu_\ell',v_\ell'+\ell\eta}$. Then $f_\eta$ alternates in sign along
$x_0<\dots<x_S$, so it has at least $S$ real zeros, and \cref{hn:fact-kmv} gives
$S\le2(L'-1)\le2(L-1)$.
\end{remark}

The same numerical bound is attained by the following repeated-variance example
\citep[proof of Proposition~7]{kalai2010efficiently}. Its upper bound needs a separate argument,
because the narrow components have equal variances.

\begin{proposition}[Repeated narrow variances attain the zero bound]
\label{hn:prop-equalnarrow-zeros}
For every integer $\Mhn\ge2$, the function
$\Mhn\hnphi_{0,\Mhn^2}-\sum_{i=1}^{\Mhn-1}\hnphi_{i,1/25}$ has exactly $2(\Mhn-1)$
distinct real zeros, all of which are simple.
\end{proposition}

\begin{proof}
Put $m=\Mhn-1$ and denote the displayed function by $f_\Mhn$. We first bound its zeros
counted with multiplicity, then exhibit alternating signs at $2m+1$ points.

Multiplication by a strictly positive factor preserves all real zeros and their multiplicities.
Expanding the Gaussian densities therefore reduces the upper bound to
\[
F_\Mhn(x):=\sqrt{2\pi}e^{25x^2/2}f_\Mhn(x)
 =e^{ax^2}-5\sum_{i=1}^me^{25ix-25i^2/2},\qquad
a=\frac{25-\Mhn^{-2}}2>0.
\]
Let $D$ denote differentiation with respect to $x$ and define
\[
L_r=\prod_{i=1}^r(D-25i)\quad(1\le r\le m),\qquad L_0=\mathrm{Id}.
\]
The factors commute, and $D-25i$ annihilates $e^{25ix-25i^2/2}$. Hence $L_m$ annihilates the
whole narrow-exponential sum. Write
\[
L_re^{ax^2}=P_r(x)e^{ax^2}.
\]
Starting from $P_0=1$, differentiation gives
\[
P_{r+1}=P_r'+\bigl(2ax-25(r+1)\bigr)P_r\qquad(0\le r<m).
\]
Induction shows that $P_r$ has degree exactly $r$ and leading coefficient $(2a)^r$, because
$2axP_r$ is the only term of degree $r+1$. Consequently
\[
L_mF_\Mhn=P_m(x)e^{ax^2}
\]
has at most $m$ real zeros counted with multiplicity and is not identically zero.

For every real $b$ and analytic function $G$,
\[
(D-b)G=e^{bx}D(e^{-bx}G).
\]
The positive exponential factors preserve zeros and multiplicities. A finite collection of
zeros of $G$ with total multiplicity $N$ gives at least $N-1$ zeros, counted with multiplicity,
of $(D-b)G$: differentiation lowers each selected multiplicity by one, and Rolle's theorem
supplies a zero between successive selected points. Every intermediate function obtained by
applying the factors of $L_m$ to $F_\Mhn$ is nonzero, since its final image $P_me^{ax^2}$ is
nonzero. Iterating the multiplicity bound through the $m$ factors gives at least $N-m$ zeros
of $L_mF_\Mhn$. Thus $N-m\le m$, or $N\le2m$. The bound holds for every finite selection,
so it also excludes an infinite real zero set. This proves the upper bound $2m$ for
$F_\Mhn$ and hence for $f_\Mhn$, with multiplicities counted.

For the lower bound, use the positive constant normalization
\[
H_\Mhn(x):=\sqrt{2\pi}f_\Mhn(x)
 =e^{-x^2/(2\Mhn^2)}-5\sum_{i=1}^me^{-(25/2)(x-i)^2}.
\]
At every integer $i=1,\dots,m$, the broad term is at most one and the matching narrow term
equals five, so $H_\Mhn(i)\le1-5<0$. At each half-integer $x_j=j+1/2$, $0\le j\le m$, we
have $0<x_j<\Mhn$, and the broad term is strictly greater than $e^{-1/2}$.

For each nonnegative integer $r$, at most two of the centers $i\in\{1,\dots,m\}$ have
$|x_j-i|=r+1/2$. Since
\[
(r+1/2)^2=1/4+r(r+1)\ge1/4+2r,
\]
the entire narrow contribution is bounded by
\[
\begin{aligned}
5\sum_{i=1}^me^{-(25/2)(x_j-i)^2}
&\le10\sum_{r=0}^{\infty}e^{-(25/2)(r+1/2)^2}\\
&\le10e^{-25/8}\sum_{r=0}^{\infty}e^{-25r}
 =\frac{10e^{-25/8}}{1-e^{-25}}.
\end{aligned}
\]
This quantity is strictly less than $e^{-1/2}$. Indeed, the positive exponential series gives
\[
e^{21/8}>e^{5/2}>\sum_{r=0}^{5}\frac{(5/2)^r}{r!}
 =\frac{8963}{768}>11,
\]
while $e^{25}>1+25=26>11$ implies $1-e^{-25}>10/11$. Hence
\[
e^{21/8}(1-e^{-25})>11\frac{10}{11}=10,
\]
which is exactly the required strict comparison. Thus $H_\Mhn(x_j)>0$ at every selected
half-integer.

The values along
\[
\frac12,\ 1,\ \frac32,\ 2,\ \dots,\ m,\ m+\frac12
\]
alternate strictly in sign. Each of the $2m$ disjoint open intervals between consecutive
points contains a real zero by the intermediate value theorem. The upper bound counts
multiplicities, so these are all the real zeros, exactly one in each interval, and every zero
is simple. Since $m=\Mhn-1$, their number is $2(\Mhn-1)$.
\end{proof}

\hnsubsection{Moment form and proof of the theorem}\label{hn:subsec-proofmain}

\begin{proposition}[A mixture of $\Mhn$ Gaussians cannot match $2\Mhn+1$ moments of a Gaussian]
\label{hn:prop-moment}
Let $\Mhn\ge1$, $p_j>0$ and $(a_j,v_j)\in\RR\times(0,\infty)$ for $j\le\Mhn$, $P=\sum_jp_j$ and
$\bar v>0$. If $\mu=\sum_jp_j\Normal(a_j,v_j)$ satisfies $\hnmom_k(\mu)=P\,\hnmom_k(\Normal(0,\bar v))$ for
$k=0,1,\dots,2\Mhn$, then $(a_j,v_j)=(0,\bar v)$ for every $j$. The range $k\le2\Mhn$ cannot be
shortened: for $0<v<\bar v$ and the rule $(x_j,\pi_j)$ of \cref{hn:lem-gh}, the mixture
$\sum_j\pi_j\Normal(\sqrt{\bar v-v}\,x_j,v)$ has the moments of $\Normal(0,\bar v)$ of orders
$0,\dots,2\Mhn-1$.
\end{proposition}

\begin{proof}
Let $f=\sum_jp_j\hnphi_{a_j,v_j}-P\hnphi_{0,\bar v}$, a combination of $L=\Mhn+1$ Gaussian densities
with finite absolute moments of all orders. By \cref{hn:lem-zerocount}, $\hnsc(f)\le2\Mhn$, while
its moments of orders $0,\dots,2\Mhn$ vanish; so \cref{hn:lem-signchange} forces $f\equiv0$. Merge
equal pairs among the $(a_j,v_j)$: each distinct pair receives a coefficient that is a sum of some
$p_j$, hence positive. If some distinct pair differed from $(0,\bar v)$, \cref{hn:lem-linind}, applied
to the distinct pairs together with $(0,\bar v)$, would make its positive coefficient vanish. Hence
every $(a_j,v_j)$ equals $(0,\bar v)$. For the last claim, the rule $(\sqrt{\bar v-v}\,x_j,\pi_j)$ has the
moments of $\Normal(0,\bar v-v)$ up to order $2\Mhn-1$, and convolution with $\Normal(0,v)$ acts on
moment sequences by an invertible triangular map with unit diagonal,
$\hnmom_n(\nu*\Normal(0,v))=\sum_{k\le n}\binom nk\hnmom_k(\nu)\,\hnmom_{n-k}(\Normal(0,v))$.
\end{proof}

\begin{proof}[Proof of \cref{hn:thm-main}]
Let $\Mhn\ge2$.

\emph{Existence for $r\le2\Mhn-1$.} By \cref{hn:prop-witness}, the Gauss--Hermite witness is a
non-trivial real solution of $\hnsys{\Mhn}{2\Mhn-1}$, hence of $\hnsys{\Mhn}{r}$ for every
$r\le2\Mhn-1$.

\emph{Rigidity at order $2\Mhn$.} Let $(c,a,b)$ be a real solution of $\hnsys{\Mhn}{2\Mhn}$ with all
$c_j\neq0$ (this part does not use $\Mhn\ge2$). Put $p_j=c_j^2>0$, $P=\sum_jp_j$,
$\bar v=1+2\max_j|b_j|$ and $v_j=\bar v+2b_j\ge1$. By \cref{hn:lem-twist}, $\mu=\sum_jp_j\Normal(a_j,v_j)$ has
the moments of $P\,\Normal(0,\bar v)$ of orders $0,\dots,2\Mhn$. By \cref{hn:prop-moment}, $a_j=0$ and
$v_j=\bar v$, that is, $b_j=0$, for all $j$.

\emph{Non-existence for $r\ge2\Mhn$.} A real solution of $\hnsys{\Mhn}{r}$ with all $c_j\neq0$
solves $\hnsys{\Mhn}{2\Mhn}$, so all $a_j=0$ and it is trivial. Hence the least $r$ without a
non-trivial real solution is $2\Mhn$, and $\rbar(k-k_0)=2(k-k_0+1)$ with $\Mhn=k-k_0+1$.
\end{proof}

\paragraph{Role of the hypotheses.} The assumption that $(a_j,b_j)$ are real ensures that
the shifted measure $\mu$ in Lemma~\ref{hn:lem-twist} is a genuine Gaussian mixture. Since each $c_j$ is real and non-zero, its weight is $p_j=c_j^2>0$; this
positivity prevents cancellation between identical components in \cref{hn:prop-moment}. \Cref{hn:rem-hypotheses} exhibits failure under joint
complexification of $(a_j,b_j,c_j)$, with signed weights, or when zero weights are allowed.

\hnsubsection{Fixed-weight arcs and contact order}\label{hn:subsec-arcs}

We next read $\hnsys{\Mhn}{r}$ as cancellation of the initial density terms along a
fixed-weight family. This connects the polynomial equations to the leading density change
along the prescribed path. Fix
$\theta_0\in\RR$, $\sigma_0>0$, weights $p_j>0$ with $\sum_jp_j=1$, and $(a_j,b_j)\in\RR^2$. For
$t\in\RR$ with $2|b_j|t^2<\sigma_0^2$ for all $j$ let
\begin{equation}\label{hn:eq-arc}
G_t=\sum_{j=1}^{\Mhn}p_j\,\delta_{(\theta_0+a_jt,\ \sigma_0^2+2b_jt^2)},\qquad
G_0=\delta_{(\theta_0,\sigma_0^2)} ,
\end{equation}
a \emph{fixed-weight arc}: locations move at speed $t$, variances at speed $t^2$. Put
$D_\alpha(x)=\partial_\theta^\alpha\hnphi_{\theta,\sigma_0^2}(x)|_{\theta=\theta_0}
=\sigma_0^{-\alpha}\hnHe_\alpha\big(\tfrac{x-\theta_0}{\sigma_0}\big)\hnphi_{\theta_0,\sigma_0^2}(x)$,
and let $\hnE_\alpha=\sum_jp_j\hne_\alpha(a_j,b_j)$ as in \eqref{hn:eq-Ealpha} with $c_j=\sqrt{p_j}$.

\begin{proposition}[Contact order of fixed-weight arcs]\label{hn:prop-arc}
\begin{enumerate}[label=(\roman*)]
\item If $2|b_j|t^2<\sigma_0^2$ for all $j$, then for every $x\in\RR$
\[
p_{G_t}(x)-p_{G_0}(x)=\sum_{\alpha\ge1}t^\alpha\hnE_\alpha D_\alpha(x),
\]
the series converging absolutely and uniformly in $x$.
\item Let $\hnord=\min\{\alpha\ge1:\hnE_\alpha\neq0\}\in\NN\cup\{\infty\}$. If $\hnord=\infty$, then
$a=b=0$ and $G_t=G_0$. If $\hnord<\infty$, then as $t\to0$
\[
\Big\|p_{G_t}-p_{G_0}-t^{\hnord}\hnE_{\hnord}D_{\hnord}\Big\|_{L^1}=O(|t|^{\hnord+1}),\qquad
\TV(p_{G_t},p_{G_0})=\frac{|\hnE_{\hnord}|\,\EE|\hnHe_{\hnord}(Z)|}{2\sigma_0^{\hnord}}|t|^{\hnord}
+O(|t|^{\hnord+1}),
\]
and the first estimate also holds in the supremum norm.
\item Consequently $(\sqrt{p_j},a_j,b_j)_j$ solves $\hnsys{\Mhn}{r}$ if and only if
$\hnord\ge r+1$, that is, if and only if $\TV(p_{G_t},p_{G_0})=O(|t|^{r+1})$. Every fixed-weight
arc with some $a_j\neq0$ has $\hnord\le2\Mhn$, and the Gauss--Hermite arc ($p_j=\pi_j$,
$a_j=x_j$, $b_j=-\frac12$) has $\hnord=2\Mhn$.
\end{enumerate}
\end{proposition}

\begin{proof}
(i) By Fourier inversion,
$\hnphi_{\theta_0+at,\sigma_0^2+2bt^2}(x)=\frac1{2\pi}\int e^{i\xi(x-\theta_0)-\sigma_0^2\xi^2/2}\,
e^{a(-i\xi t)+b(-i\xi t)^2}\,\dd\xi$. By \eqref{hn:eq-genfun} the second exponential equals
$\sum_\alpha\hne_\alpha(a,b)(-i\xi)^\alpha t^\alpha$, and
$\sum_\alpha|\hne_\alpha(a,b)||\xi t|^\alpha\le\sum_\alpha\hne_\alpha(|a|,|b|)|\xi t|^\alpha
=e^{|a\xi t|+|b|\xi^2t^2}$, which is integrable against $e^{-\sigma_0^2\xi^2/2}$ when
$2|b|t^2<\sigma_0^2$. Dominated convergence allows termwise integration, uniformly in $x$, and
$\frac1{2\pi}\int(-i\xi)^\alpha e^{i\xi(x-\theta_0)-\sigma_0^2\xi^2/2}\,\dd\xi=(-\partial_x)^\alpha
\hnphi_{\theta_0,\sigma_0^2}(x)=D_\alpha(x)$. Summing over $j$ with weights $p_j$, the term
$\alpha=0$ is $p_{G_0}$.

(ii) If $\hnord=\infty$, then $(\sqrt{p_j},a_j,b_j)$ solves $\hnsys{\Mhn}{r}$ for every $r$, and the
rigidity part of \cref{hn:thm-main} gives $a=b=0$. Let $\hnord<\infty$. We first show that
$\sum_\alpha\sqrt{\alpha!}\,\hne_\alpha(|a|,|b|)\,\tau^\alpha<\infty$ whenever $4|b|\tau^2<1$.
Since $\alpha!\le2^\alpha n_1!\,(2n_2)!\le2^\alpha n_1!\,4^{n_2}(n_2!)^2$ for $\alpha=n_1+2n_2$,
\[
\sqrt{\alpha!}\,\frac{|a|^{n_1}|b|^{n_2}\tau^\alpha}{n_1!\,n_2!}
\le\frac{(\sqrt2|a|\tau)^{n_1}}{\sqrt{n_1!}}\,(4|b|\tau^2)^{n_2},
\]
and the double series over $(n_1,n_2)$ converges. Next,
$\|D_\alpha\|_{L^1}=\sigma_0^{-\alpha}\EE|\hnHe_\alpha(Z)|\le\sigma_0^{-\alpha}\sqrt{\alpha!}$ by
\cref{hn:lem-gh}(i) and Cauchy--Schwarz. Fix $t_1>0$ with $4|b_j|t_1^2<\sigma_0^2$ for all $j$.
For $|t|\le t_1$ the series in (i) converges absolutely in $L^1$, so its $L^1$-sum equals the
pointwise sum, and
\[
\Big\|\sum_{\alpha>\hnord}t^\alpha\hnE_\alpha D_\alpha\Big\|_{L^1}
\le|t|^{\hnord+1}\sum_{\alpha>\hnord}\Big(\sum_jp_j\hne_\alpha(|a_j|,|b_j|)\Big)
\sqrt{\alpha!}\,\sigma_0^{-\alpha}t_1^{\alpha-\hnord-1}=O(|t|^{\hnord+1}).
\]
This is the $L^1$ estimate; the formula for $\TV=\frac12\|\cdot\|_{L^1}$ follows because
$\|D_\hnord\|_{L^1}=\sigma_0^{-\hnord}\EE|\hnHe_\hnord(Z)|$. For the supremum norm fix $t_2>0$ with
$2|b_j|t_2^2<\sigma_0^2$ for all $j$. Since
$|D_\alpha(x)|\le\frac1{2\pi}\int|\xi|^\alpha e^{-\sigma_0^2\xi^2/2}\dd\xi$,
$|\hnE_\alpha|\le\sum_jp_j\hne_\alpha(|a_j|,|b_j|)$ and $|t|^\alpha\le|t|^{\hnord+1}t_2^{\alpha-\hnord-1}$ for
$\alpha>\hnord$ and $|t|\le t_2$, summing the exponential series as in (i) gives
\[
\sup_x\Big|\sum_{\alpha>\hnord}t^\alpha\hnE_\alpha D_\alpha(x)\Big|
\le\frac{|t|^{\hnord+1}}{t_2^{\hnord+1}}\,\frac1{2\pi}\int\sum_jp_j\,
e^{|a_j\xi|t_2+|b_j|\xi^2t_2^2}\,e^{-\sigma_0^2\xi^2/2}\,\dd\xi=O(|t|^{\hnord+1}).
\]

(iii) The first claim is the definition of $\hnord$ together with (ii), since
$\EE|\hnHe_\hnord(Z)|>0$. If some $a_j\neq0$, the rigidity part of \cref{hn:thm-main} excludes
$\hnord\ge2\Mhn+1$; for the Gauss--Hermite arc, $\hnord=2\Mhn$ by \cref{hn:prop-witness}.
\end{proof}

For the minimax lower bounds in Appendix~\ref{hn:sec-stat}, we supplement the density-contact
asymptotics with an exact Hermite expansion of the chi-square divergence.

\begin{lemma}[Hermite expansion at the finite chi-square boundary]\label{hn:lem-chi2}
Let
\[
G=\sum_{j=1}^Lp_j\delta_{(a_j,v_j)}
\]
be a probability measure on $\RR\times(0,\infty)$ with $p_j>0$, and put
\[
\hnE_\alpha(G)=\sum_jp_j\hne_\alpha\!\left(a_j,\frac{v_j-1}{2}\right).
\]
Then the following conditions are equivalent:
\begin{enumerate}[label=(\roman*)]
\item $0<v_j<2$ for every $j$;
\item $\chisq(p_G\|\hnphi)<\infty$.
\end{enumerate}
When these conditions hold,
\[
\frac{p_G}{\hnphi}=\sum_{\alpha\ge0}\hnE_\alpha(G)\hnHe_\alpha
\]
pointwise on $\RR$ and absolutely in $L^2(\hnphi(x)\,\dd x)$, in the sense that
\[
\sum_{\alpha\ge0}\sqrt{\alpha!}\,|\hnE_\alpha(G)|<\infty.
\]
Consequently,
\[
\chisq(p_G\|\hnphi)
 =\int_{\RR}\frac{(p_G-\hnphi)^2}{\hnphi}\,\dd x
 =\sum_{\alpha\ge1}\alpha!\,\hnE_\alpha(G)^2.
\]
In particular, one positive-weight component with variance $v_j\ge2$, including $v_j=2$,
makes the chi-square divergence infinite.

Along the Gauss--Hermite arc
$G_t=\sum_j\pi_j\delta_{(x_jt,\,1-t^2)}$, $0<|t|\le\frac12$,
\[
\chisq(p_{G_t}\|\hnphi)
 =t^{4\Mhn}\sum_{\alpha\ge2\Mhn}
 \alpha!\,\bigl(\hnE_\alpha^{\mathrm{GH}}\bigr)^2t^{2(\alpha-2\Mhn)}
 \le C_\Mhn |t|^{4\Mhn},
\]
where
\[
\hnE_\alpha^{\mathrm{GH}}
 =\sum_j\pi_j\hne_\alpha(x_j,-1/2),
\qquad
C_\Mhn=\sum_{\alpha\ge2\Mhn}
 \alpha!\,\bigl(\hnE_\alpha^{\mathrm{GH}}\bigr)^2
 4^{-(\alpha-2\Mhn)}<\infty.
\]
Moreover,
\[
\lim_{t\to0}\frac{\chisq(p_{G_t}\|\hnphi)}{t^{4\Mhn}}
 =\frac{(\Mhn!)^2}{(2\Mhn)!}.
\]
\end{lemma}

\begin{proof}
We first record the required summability estimate. For non-negative integers $m,\ell$ and
$\delta>0$,
\[
\binom{m+2\ell}{m}
 \le(1+\delta)^m(1+\delta^{-1})^{2\ell},
\qquad
\frac{\sqrt{(2\ell)!}}{\ell!}
 =\sqrt{\binom{2\ell}{\ell}}\le2^\ell.
\]
It follows that
\begin{equation}\label{hn:eq-factorial-bound}
\frac{\sqrt{(m+2\ell)!}}{m!\ell!}
 \le\frac{(1+\delta)^{m/2}}{\sqrt{m!}}\,
 \bigl[2(1+\delta^{-1})\bigr]^\ell.
\end{equation}
Fix $a,b\in\RR$ with $B:=|b|<\frac12$. Choose $\delta>0$ so that
$r_B:=2B(1+\delta^{-1})<1$; when $B=0$, any $\delta>0$ is admissible. With
$A=|a|\sqrt{1+\delta}$, equation~\eqref{hn:eq-factorial-bound} gives
\[
\begin{aligned}
\sum_{\alpha\ge0}\sqrt{\alpha!}\,|\hne_\alpha(a,b)|
&\le\sum_{m,\ell\ge0}
 \frac{\sqrt{(m+2\ell)!}\,|a|^mB^\ell}{m!\ell!}\\
&\le
 \left(\sum_{m\ge0}\frac{A^m}{\sqrt{m!}}\right)
 \left(\sum_{\ell\ge0}r_B^\ell\right)<\infty.
\end{aligned}
\]
The first series converges because the ratio of consecutive terms tends to zero, and the second
is geometric.

We next prove the pointwise expansion. Fourier inversion gives
\[
\hnphi_{a,v}(x)=\frac1{2\pi}\int_{\RR}
 e^{i\xi x-\xi^2/2}
 \exp\!\left(a(-i\xi)+\frac{v-1}{2}(-i\xi)^2\right)\,\dd\xi.
\]
If $0<v<2$, then $b=(v-1)/2$ satisfies $|b|<1/2$. After the second exponential is expanded,
the sum of the absolute values of its terms is bounded by
$\exp(|a\xi|+|b|\xi^2)$. Multiplication by $e^{-\xi^2/2}$ leaves an integrable function.
Dominated convergence therefore permits termwise Fourier inversion, uniformly in $x$. Since
\[
\frac1{2\pi}\int_{\RR}e^{i\xi x-\xi^2/2}(-i\xi)^\alpha\,\dd\xi
 =\left(-\frac{\dd}{\dd x}\right)^\alpha\hnphi(x)
 =\hnHe_\alpha(x)\hnphi(x),
\]
we obtain
\[
\hnphi_{a,v}(x)
 =\hnphi(x)\sum_{\alpha\ge0}
 \hne_\alpha\!\left(a,\frac{v-1}{2}\right)\hnHe_\alpha(x).
\]
Summing the finitely many component expansions gives the asserted pointwise series for
$p_G/\hnphi$.

Repeated integration by parts, as in Lemma~\ref{hn:lem-gh}(i), gives
\[
\int_{\RR}\hnHe_\alpha(x)\hnHe_\beta(x)\hnphi(x)\,\dd x
 =\alpha!\,\one\{\alpha=\beta\}.
\]
Indeed, if $\alpha>\beta$ the $\alpha$-th derivative of $\hnHe_\beta$ vanishes; if
$\alpha<\beta$ one interchanges the two polynomials; and equality gives $\alpha!$ because
$\hnHe_\alpha$ is monic. Thus
$\|\hnHe_\alpha\|_{L^2(\hnphi\,\dd x)}=\sqrt{\alpha!}$.
The factorial estimate and the triangle inequality now give
\[
\sum_{\alpha\ge0}\sqrt{\alpha!}\,|\hnE_\alpha(G)|
 \le\sum_jp_j\sum_{\alpha\ge0}
 \sqrt{\alpha!}\left|
 \hne_\alpha\!\left(a_j,\frac{v_j-1}{2}\right)\right|<\infty.
\]
Hence the Hermite series converges absolutely in the Banach space
$L^2(\hnphi\,\dd x)$; denote its limit by $S$.

We identify this limit directly by Borel--Cantelli.
Choose a subsequence of partial sums $S_{N_k}$ with
$\|S_{N_k}-S\|_2^2\le2^{-3k}$. Markov's inequality gives
\[
\hnphi(x)\,\dd x\{\,|S_{N_k}-S|>2^{-k}\,\}\le2^{-k}.
\]
The sum of these measures is finite, so $S_{N_k}\to S$ almost everywhere. The full sequence
converges pointwise to $p_G/\hnphi$, and therefore $S=p_G/\hnphi$ almost everywhere.
Orthogonality yields
\[
\int_{\RR}\frac{p_G^2}{\hnphi}\,\dd x
 =\sum_{\alpha\ge0}\alpha!\,\hnE_\alpha(G)^2.
\]
Since $\hnE_0(G)=1$, subtracting the constant mode proves the chi-square identity and its
finiteness under condition (i).

It remains to prove the converse. Direct calculation gives, for a positive constant $D_{a,v}$,
\[
\frac{\hnphi_{a,v}(x)^2}{\hnphi(x)}
 =D_{a,v}\exp\!\left[
 \left(\frac12-\frac1v\right)x^2+\frac{2a}{v}x\right].
\]
If $v>2$, the coefficient of $x^2$ is positive, so this function is not integrable. If $v=2$,
it is either a positive constant or a positive constant times a non-constant real exponential,
and again is not integrable. Because every $p_j$ is positive,
$p_G\ge p_j\hnphi_{a_j,v_j}$; hence one component with $v_j\ge2$ forces
$\int p_G^2/\hnphi=\infty$. On the other hand,
\[
p_G^2\le2(p_G-\hnphi)^2+2\hnphi^2,
\]
so finite chi-square divergence would imply $\int p_G^2/\hnphi<\infty$. This contradiction
proves the equivalence, including the boundary $v_j=2$.

For the Gauss--Hermite arc, homogeneity gives
\[
\hne_\alpha(tx_j,-t^2/2)=t^\alpha\hne_\alpha(x_j,-1/2),
\qquad
\hnE_\alpha(G_t)=t^\alpha\hnE_\alpha^{\mathrm{GH}}.
\]
Proposition~\ref{hn:prop-witness} gives
$\hnE_\alpha^{\mathrm{GH}}=0$ for $1\le\alpha<2\Mhn$ and
$\hnE_{2\Mhn}^{\mathrm{GH}}=-\Mhn!/(2\Mhn)!$. The chi-square identity therefore gives the
displayed series. At $t=\frac12$ it also gives
\[
C_\Mhn=4^{2\Mhn}\chisq(p_{G_{1/2}}\|\hnphi)<\infty.
\]
For $0<|t|\le\frac12$,
$|t|^{2(\alpha-2\Mhn)}\le4^{-(\alpha-2\Mhn)}$, which proves the upper bound. The same summable
majorant permits termwise passage to the limit as $t\to0$, and the surviving term is
\[
(2\Mhn)!\left(\frac{\Mhn!}{(2\Mhn)!}\right)^2
 =\frac{(\Mhn!)^2}{(2\Mhn)!}.
\]
\end{proof}

The identity in \cref{hn:lem-chi2} is the location--scale form of the Hermite expansion of the
$\chisq$-divergence used for Gaussian location mixtures, for instance by \citet{wu2020dmm}.

\begin{lemma}[Total-variation contact of the Gauss--Hermite arc]
\label{hn:lem-gh-tv}
For the arc in Lemma~\ref{hn:lem-chi2},
\[
\TV(p_{G_t},\hnphi)
 =\kappa_\Mhn|t|^{2\Mhn}+O(|t|^{2\Mhn+1}),
\qquad
\kappa_\Mhn=
 \frac{\Mhn!}{2(2\Mhn)!}
 \int_{\RR}|\hnHe_{2\Mhn}(z)|\hnphi(z)\,\dd z>0.
\]
\end{lemma}

\begin{proof}
Apply the absolute $L^2$ summability in Lemma~\ref{hn:lem-chi2} at a fixed
$\rho\in(0,\frac12]$. Since the coefficient of $\hnHe_\alpha$ for $G_\rho$ is
$\rho^\alpha\hnE_\alpha^{\mathrm{GH}}$,
\[
\sum_{\alpha\ge0}\rho^\alpha\sqrt{\alpha!}\,
 |\hnE_\alpha^{\mathrm{GH}}|<\infty.
\]
Cauchy--Schwarz and Hermite orthogonality give
\[
\|\hnHe_\alpha\hnphi\|_{L^1}
 \le\sqrt{\alpha!}.
\]
For $|t|\le\rho$, it follows that
\[
\begin{aligned}
&\left\|p_{G_t}-\hnphi
 -t^{2\Mhn}\hnE_{2\Mhn}^{\mathrm{GH}}\hnHe_{2\Mhn}\hnphi\right\|_1\\
&\qquad\le
 \sum_{\alpha>2\Mhn}|t|^\alpha
 |\hnE_\alpha^{\mathrm{GH}}|\,\|\hnHe_\alpha\hnphi\|_1\\
&\qquad\le
 |t|^{2\Mhn+1}\rho^{-(2\Mhn+1)}
 \sum_{\alpha>2\Mhn}\rho^\alpha\sqrt{\alpha!}\,
 |\hnE_\alpha^{\mathrm{GH}}|
 =O(|t|^{2\Mhn+1}).
\end{aligned}
\]
The reverse triangle inequality for the $L^1$ norm and
$\hnE_{2\Mhn}^{\mathrm{GH}}=-\Mhn!/(2\Mhn)!$ give the formula. The constant is positive because
$\hnHe_{2\Mhn}$ is a non-zero polynomial and $\hnphi$ is everywhere positive.
\end{proof}

\begin{proposition}[Softmax descendants have exact Hellinger contact]
\label{hn:prop-softmax-contact}
Fix $\Mhn\ge2$ and a scalar covariate $u$. For expert parameters
$(\beta_0,\beta_1,a,b,v)$, use the conditional-density convention
\[
g_{\mathcal A}(y\mid u)=
 \frac{\sum_j e^{\beta_{0j}+\beta_{1j}u}
 \hnphi_{a_ju+b_j,v_j}(y)}
 {\sum_j e^{\beta_{0j}+\beta_{1j}u}}.
\]
Let $U\sim\mathrm{Unif}[-1,1]$. Take the two true experts
\[
(0,1,1,0,1),\qquad (0,0,-1,1,1).
\]
For $|t|<1$, replace the first expert by the $\Mhn$ descendants
\[
(\log\pi_j,1,1,tx_j,1-t^2),\qquad 1\le j\le\Mhn,
\]
and leave the second expert unchanged. Denote the true and fitted conditional densities by
$g_*$ and $g_t$, and put
\[
w(u)=\frac{e^u}{1+e^u},
\qquad
q_t=\sum_{j=1}^{\Mhn}\pi_j\hnphi_{tx_j,1-t^2}.
\]
Then
\[
g_*(y\mid u)
 =w(u)\hnphi(y-u)+(1-w(u))\hnphi(y+u-1),
\]
\[
g_t(y\mid u)
 =w(u)q_t(y-u)+(1-w(u))\hnphi(y+u-1).
\]
Let
\[
w_-=\frac1{1+e},\qquad w_+=\frac e{1+e}.
\]
There exists $t_\Mhn\in(0,\frac12]$ such that, for every $u\in[-1,1]$ and
$0<|t|\le t_\Mhn$,
\[
\frac{w_-^2\kappa_\Mhn^2}{8}|t|^{4\Mhn}
 \le\Hel^2(g_t(\cdot\mid u),g_*(\cdot\mid u))
 \le\frac{w_+C_\Mhn}{2}|t|^{4\Mhn}.
\]
Consequently,
\[
\EE_U\Hel^2(g_t(\cdot\mid U),g_*(\cdot\mid U))
 \asymp_\Mhn |t|^{4\Mhn}.
\]
The same expectation equals the squared Hellinger distance between the fitted and true joint
densities when they use the same uniform covariate density.
\end{proposition}

\begin{proof}
The gate numerator of descendant $j$ is
$e^{\log\pi_j+u}=\pi_je^u$. Their sum is $e^u\sum_j\pi_j=e^u$, while the unchanged second gate
numerator is $1$. Thus the fitted and true denominators both equal $1+e^u$. The descendant means
are $u+tx_j$, so their normalized cluster is
\[
\sum_j\pi_j\hnphi_{u+tx_j,1-t^2}(y)=q_t(y-u).
\]
The unchanged second mean is $-u+1$. This proves the two displayed formulas and the identity
\begin{equation}\label{hn:eq-softmax-contact-difference}
g_t(y\mid u)-g_*(y\mid u)
 =w(u)\{q_t(y-u)-\hnphi(y-u)\}.
\end{equation}

We use two elementary Hellinger comparisons. For arbitrary probability densities $p,q$,
Cauchy--Schwarz gives
\[
\begin{aligned}
2\TV(p,q)
&=\int|\sqrt p-\sqrt q|(\sqrt p+\sqrt q)\\
&\le
 \left(\int(\sqrt p-\sqrt q)^2\right)^{1/2}
 \left(\int(\sqrt p+\sqrt q)^2\right)^{1/2}
 \le2\sqrt{2\Hel^2(p,q)}.
\end{aligned}
\]
Hence
\begin{equation}\label{hn:eq-hellinger-tv-lower}
\Hel^2(p,q)\ge\frac12\TV(p,q)^2.
\end{equation}
Translation invariance and equation~\eqref{hn:eq-softmax-contact-difference} give
\[
\TV(g_t(\cdot\mid u),g_*(\cdot\mid u))
 =w(u)\TV(q_t,\hnphi).
\]
For the upper bound, use
\[
(\sqrt p-\sqrt q)^2=\frac{(p-q)^2}{(\sqrt p+\sqrt q)^2}.
\]
Since $g_*(y\mid u)\ge w(u)\hnphi(y-u)$,
\[
\begin{aligned}
\Hel^2(g_t(\cdot\mid u),g_*(\cdot\mid u))
&\le\frac12\int
 \frac{w(u)^2\{q_t(y-u)-\hnphi(y-u)\}^2}
      {w(u)\hnphi(y-u)}\,\dd y\\
&=\frac{w(u)}2\chisq(q_t\|\hnphi).
\end{aligned}
\]

For $u\in[-1,1]$, the monotonicity of the logistic function gives
$w_-\le w(u)\le w_+$. By Lemma~\ref{hn:lem-gh-tv}, after decreasing
$t_\Mhn\le\frac12$ if necessary,
\[
\TV(q_t,\hnphi)\ge\frac{\kappa_\Mhn}{2}|t|^{2\Mhn}
\qquad(0<|t|\le t_\Mhn).
\]
Combining this estimate with equation~\eqref{hn:eq-hellinger-tv-lower} gives the stated lower
bound. Lemma~\ref{hn:lem-chi2} gives the stated upper bound. Both estimates are uniform in
$u\in[-1,1]$, so integration against the law of $U$ preserves them.

Finally, if $f_U$ is the common covariate density, direct substitution into the Hellinger
definition gives
\[
\frac12\iint
 \left(\sqrt{f_U(u)g_t(y\mid u)}
       -\sqrt{f_U(u)g_*(y\mid u)}\right)^2\,\dd y\,\dd u
 =\EE_U\Hel^2(g_t(\cdot\mid U),g_*(\cdot\mid U)).
\]
\end{proof}

The local asymptotics as $t\to0$ combine with continuity on compact subintervals of
$(0,1/2]$ to give sharpness along the whole Gauss--Hermite path.

\begin{proposition}[Sharpness on the continuous Gauss--Hermite path]
\label{hn:prop-gh-sharpness}
Let $\Mhn\ge2$ and
\[
G_t=\sum_{j=1}^{\Mhn}\pi_j\delta_{(tx_j,1-t^2)},
\qquad 0<t\le\frac12,
\qquad
G_0=\delta_{(0,1)}.
\]
For every real $r>0$,
\[
\inf_{0<t\le1/2}
\frac{\TV(p_{G_t},p_{G_0})}
     {\bigl(W_1^{\mathrm E}(G_t,G_0)\bigr)^r}>0
\quad\Longleftrightarrow\quad r\ge2\Mhn.
\]
At the boundary,
\[
\lim_{t\downarrow0}
\frac{\TV(p_{G_t},p_{G_0})}
     {\bigl(W_1^{\mathrm E}(G_t,G_0)\bigr)^{2\Mhn}}
 =\frac{\kappa_\Mhn}{c_{\mathrm E}^{\,2\Mhn}},
\qquad
c_{\mathrm E}=\sum_{j=1}^{\Mhn}\pi_j|x_j|>0.
\]
\end{proposition}

\begin{proof}
Because the second marginal is the point mass $G_0$, every coupling sends each atom of $G_t$ to
$(0,1)$. Hence
\[
W_1^{\mathrm E}(G_t,G_0)
 =\sum_j\pi_j\sqrt{t^2x_j^2+t^4}
 =t\sum_j\pi_j\sqrt{x_j^2+t^2},
\]
and therefore
\[
\frac{W_1^{\mathrm E}(G_t,G_0)}t\longrightarrow
c_{\mathrm E}=\sum_j\pi_j|x_j|.
\]
Quadrature exactness gives $\sum_j\pi_jx_j^2=1$, so $c_{\mathrm E}>0$.
Lemma~\ref{hn:lem-gh-tv} gives
\[
\TV(p_{G_t},p_{G_0})
 =\kappa_\Mhn t^{2\Mhn}+O(t^{2\Mhn+1}),
\qquad \kappa_\Mhn>0.
\]

The numerator is strictly positive for every $0<t\le\frac12$. Indeed,
Lemma~\ref{hn:lem-chi2} gives
\[
\chisq(p_{G_t}\|\hnphi)
 =\sum_{\alpha\ge2\Mhn}
 \alpha!\,\bigl(\hnE_\alpha^{\mathrm{GH}}\bigr)^2t^{2\alpha},
\]
whose $\alpha=2\Mhn$ summand is positive. Thus $p_{G_t}$ cannot equal $\hnphi$ almost
everywhere.

It remains to control the path away from zero. For $0\le t\le\frac12$, all component means lie
in a fixed bounded interval and all variances lie in $[3/4,1]$. If
$A=\frac12\max_j|x_j|$, then
\[
(z-tx_j)^2\ge\frac{z^2}{2}-A^2.
\]
The Gaussian prefactors are uniformly bounded on this variance interval, so the component
densities are dominated by $K e^{-z^2/4}$ for a finite constant $K$, independently of $j$ and
$t$. They are pointwise continuous in $t$. Dominated convergence therefore makes
$t\mapsto p_{G_t}$ continuous in $L^1$, and hence makes
$t\mapsto\TV(p_{G_t},p_{G_0})$ continuous. The transport denominator is positive and continuous
for $t>0$.

Set
\[
F_r(t)=
\frac{\TV(p_{G_t},p_{G_0})}
     {\bigl(W_1^{\mathrm E}(G_t,G_0)\bigr)^r}.
\]
The two asymptotic formulas give
\[
F_r(t)\sim\frac{\kappa_\Mhn}{c_{\mathrm E}^{\,r}}t^{2\Mhn-r}
\qquad(t\downarrow0).
\]
If $r<2\Mhn$, then $F_r(t)\to0$. If $r=2\Mhn$, then $F_r(t)$ converges to the positive finite limit $\kappa_\Mhn/c_{\mathrm E}^{2\Mhn}$. If $r>2\Mhn$, then $F_r(t)\to+\infty$. In either of the last two cases, $F_r$ has a
positive lower bound on some interval $(0,\delta]$. On the compact interval
$[\delta,\frac12]$, it is continuous and strictly positive, so it has a positive minimum.
This proves the equivalence and the boundary limit.
\end{proof}

\hnsubsection{Escaping mixtures at fixed positive size}\label{hn:subsec-escape}

An unbounded location domain permits a different behaviour: atoms can move farther away
while their weights decrease. The construction below keeps either size constraint
fixed while the Hellinger distance tends to zero.

For a mixing measure $G=\sum_jp_j\delta_{(\theta_j,v_j)}$ relative to
$G_0=\delta_{(0,1)}$, put $\Delta\theta_j=\theta_j$ and $\Delta v_j=v_j-1$, and define
\[
\|G\|_{\mathrm h}
 =\left(\sum_jp_j\bigl(|\Delta\theta_j|^4+|\Delta v_j|^2\bigr)^{1/2}\right)^{1/2}.
\]
We write $W_1^{\mathrm E}$ for the Wasserstein distance whose ground distance on
$\RR\times(0,\infty)$ is Euclidean. The following proposition shows that neither of these fixed
positive-size constraints is coercive for Hellinger distance on an unbounded location domain.

\begin{proposition}[Escaping positive-weight mixtures]\label{prop:note-escape}
Fix $\Mhn\ge2$ and $\varepsilon>0$. Among mixing measures $G$ having exactly $\Mhn$ distinct
atoms, strictly positive weights, component variances equal to $1$, and
$\|G\|_{\mathrm h}=\varepsilon$, one has
\[
\inf_G\Hel^2(p_G,\hnphi)=0,
\qquad
\Hel^2(p,q):=\frac12\int_{\RR}(\sqrt p-\sqrt q)^2.
\]
The same conclusion holds if the homogeneous-size constraint is replaced by
$W_1^{\mathrm E}(G,\delta_{(0,1)})=\varepsilon$. In particular, the first infimum remains zero
when every variance is required to lie in $[0.1,1.9]$.
\end{proposition}

\begin{proof}
Choose distinct positive numbers $a_1,\dots,a_{\Mhn-1}$. For large $T$, put
\[
\eta_{j,T}=\frac{\varepsilon^2}{(\Mhn-1)T^2a_j^2},
\qquad
\eta_T=\sum_{j=1}^{\Mhn-1}\eta_{j,T},
\]
and define
\[
G_T=(1-\eta_T)\delta_{(0,1)}
 +\sum_{j=1}^{\Mhn-1}\eta_{j,T}\delta_{(Ta_j,1)}.
\]
For all sufficiently large $T$, every weight is positive and the measure has exactly $\Mhn$
distinct atoms. Direct substitution gives
\[
\|G_T\|_{\mathrm h}^2
 =\sum_{j=1}^{\Mhn-1}\eta_{j,T}|Ta_j|^2
 =\varepsilon^2.
\]
For non-negative $r,s$, $(\sqrt r-\sqrt s)^2\le|r-s|$. Hence
$\Hel^2(p,q)\le\TV(p,q)$ for probability densities, and therefore
\[
\Hel^2(p_{G_T},\hnphi)
 \le\TV(p_{G_T},\hnphi)
 \le\sum_{j=1}^{\Mhn-1}\eta_{j,T}
 =\frac{\varepsilon^2}{(\Mhn-1)T^2}
   \sum_{j=1}^{\Mhn-1}a_j^{-2}\longrightarrow0.
\]
This proves the homogeneous-size assertion.

For the Euclidean-transport constraint, set instead
\[
\widetilde\eta_{j,T}=\frac{\varepsilon}{(\Mhn-1)Ta_j},
\qquad
\widetilde\eta_T=\sum_{j=1}^{\Mhn-1}\widetilde\eta_{j,T},
\]
and
\[
\widetilde G_T=(1-\widetilde\eta_T)\delta_{(0,1)}
 +\sum_{j=1}^{\Mhn-1}\widetilde\eta_{j,T}\delta_{(Ta_j,1)}.
\]
A coupling with the point mass $\delta_{(0,1)}$ is unique, so
\[
W_1^{\mathrm E}(\widetilde G_T,\delta_{(0,1)})
 =\sum_{j=1}^{\Mhn-1}\widetilde\eta_{j,T}Ta_j
 =\varepsilon.
\]
The same total-variation estimate gives
\[
\Hel^2(p_{\widetilde G_T},\hnphi)
 \le\frac{\varepsilon}{(\Mhn-1)T}
   \sum_{j=1}^{\Mhn-1}a_j^{-1}\longrightarrow0.
\]
\end{proof}

\hnsection{Mixtures of experts}\label{hn:sec-moe}

Several recent analyses of Gaussian mixtures of experts reduce parameter-estimation rates to the
solvability of polynomial systems of Ho--Nguyen type, and evaluate them only for two or three
fitted experts per true expert. All of these systems are coefficient systems of one generating
function, and \cref{hn:thm-main} determines their orders. Throughout this section $\Mhn\ge2$ is the
number of fitted experts in one Voronoi cell (written $m$ in the cited papers), all unknowns are
\emph{real}, and multi-index notation is used: for $\alpha\in\NN^d$, $|\alpha|=\sum_i\alpha_i$,
$\alpha!=\prod_i\alpha_i!$ and $p^\alpha=\prod_ip_i^{\alpha_i}$.

\begin{definition}[Generating-function systems]\label{hn:def-moe}
Let $d\ge0$ be an integer. When $d=0$, the variable $u$ is absent and $F$ depends only on $v$.
The conventions $\RR^0=\{0\}$ and $\NN^0=\{0\}$ give the unique empty multi-index length $0$,
factorial $1$ and monomial $1$. This auxiliary dimension is distinct from a model's covariate dimension.
Let $\mu_1,\dots,\mu_K$ be monomials of positive total degree in $(u,v)\in\RR^d\times\RR$. For
$(w_j,\vartheta_j)\in\RR\times\RR^K$, $j\le\Mhn$, put
\[
F(u,v)=\sum_{j=1}^{\Mhn}w_j\exp\Big(\sum_{k=1}^K\vartheta_{jk}\,\mu_k(u,v)\Big).
\]
The \emph{order-$r$ system} $\mathcal S_r$ consists of the equations
$[u^{\ell_1}v^{\ell_2}]F=0$ for all $(\ell_1,\ell_2)\in\NN^d\times\NN$ with
$1\le|\ell_1|+\ell_2\le r$, where $[u^{\ell_1}v^{\ell_2}]F$ is a Taylor coefficient at the origin.
\end{definition}

The systems of interest are the following.
\begin{itemize}[leftmargin=*]
\item[(HN)] With $d=0$ and $u$ absent, let $\mu=(v,v^2)$,
$\vartheta_j=(a_j,b_j)$ and $w_j=c_j^2$. Identifying $\ell_2=\alpha$, this is \eqref{hn:eq-system}.
A solution is non-trivial if all $c_j\neq0$ and some $a_j\neq0$.
\item[(SM)] Softmax-gated Gaussian MoE, system~(9) of \citet{nguyen2023softmax}
for any $d\ge1$: $\mu=(u_1,\dots,u_d,\ u_1v,\dots,u_dv,\ v,\ v^2)$,
$\vartheta_j=(p_{1j},p_{2j},p_{3j},p_{4j})\in\RR^d\times\RR^d\times\RR\times\RR$ and $w_j=p_{5j}^2$.
The system is
$\sum_j\sum_{\alpha\in\mathcal I_{\ell_1,\ell_2}}p_{5j}^2p_{1j}^{\alpha_1}p_{2j}^{\alpha_2}
p_{3j}^{\alpha_3}p_{4j}^{\alpha_4}/(\alpha_1!\,\alpha_2!\,\alpha_3!\,\alpha_4!)=0$, with
$\mathcal I_{\ell_1,\ell_2}=\{\alpha\in\NN^d\times\NN^d\times\NN\times\NN:\alpha_1+\alpha_2=\ell_1,\
|\alpha_2|+\alpha_3+2\alpha_4=\ell_2\}$ and $1\le|\ell_1|+\ell_2\le r$. A solution is non-trivial if all
$p_{5j}\neq0$ and some $p_{3j}\neq0$. The order is denoted $\rbar_{\mathrm{SM}}(\Mhn)$ (the
$\bar r(m)$ of that paper).
\item[(GG)] Gaussian-gated MoE, $d=1$, system~(12) of \citet{nguyen2024gaussiangated}:
$\mu=(u,u^2,uv,v,v^2)$, $\vartheta_l=(q_{1l},\dots,q_{5l})$ and $w_l=p_l^2$,
with index set $\mathcal J_{\ell_1,\ell_2}=\{\alpha\in\NN^5:\alpha_1+2\alpha_2+\alpha_3=\ell_1,\
\alpha_3+\alpha_4+2\alpha_5=\ell_2\}$ and $1\le\ell_1+\ell_2\le r$. A solution is non-trivial if all $p_l\neq0$ and
some $q_{4l}\neq0$. The order is $\tilde r(\Mhn)$. System~(9) of the same paper is the scalar
(HN) system in $(p_l,q_{1l},q_{2l})$, with $u$ absent.
\item[(DS)] DeepSeekMoE with linear experts, Appendix~A of \citet{nguyen2025deepseek}:
the $r_1$-system has $u$ absent and is (HN) in
$(c,a,b)=(s_3,s_1,s_2)$; a solution is non-trivial if all $s_{3i}\neq0$ and some $s_{1i}\neq0$. The $r_2$-system
uses the positive covariate dimension $d\ge1$ and is (SM) in $(p_1,\dots,p_5)=(t_1,\dots,t_5)$,
and a solution is non-trivial if all $t_{5i}\neq0$ and some $t_{4i}\neq0$. Note that $t_4$ is the coefficient of
$v^2$, the \emph{scale} variable. For $r_2$, we restore the multi-index norm bars in the range below
Appendix~A, equation~(14), using $1\le|\ell_1|+\ell_2\le r$ in place of
$1\le\ell_1+\ell_2\le r$. This agrees with Appendix~B.2, Case~II of that paper and
equation~(9) of \citet{nguyen2023softmax}. Throughout, $r_2$ refers to this corrected range.
\end{itemize}

\begin{lemma}[Equivalence with the generating-function systems]\label{hn:lem-transcription}
In cases (HN), (SM), (GG), and (DS), expansion of the generating function gives, equation by
equation, the order-$r$ coefficient systems and index ranges of \cref{hn:def-moe}.
\end{lemma}

\begin{proof}
Write $\mu_k=u^{\gamma_k}v^{\delta_k}$. Expanding each exponential factor gives
\[
\exp\Big(\sum_k\vartheta_k\mu_k\Big)=\sum_{\beta\in\NN^K}\ \prod_k\frac{\vartheta_k^{\beta_k}\mu_k^{\beta_k}}{\beta_k!},
\]
so $[u^{\ell_1}v^{\ell_2}]F=\sum_jw_j\sum_\beta\vartheta_j^\beta/\beta!$ over all $\beta$ with
$\sum_k\beta_k\gamma_k=\ell_1$ and $\sum_k\beta_k\delta_k=\ell_2$. For (SM), split
$\beta=(\alpha_1,\alpha_2,\alpha_3,\alpha_4)$ according to the four blocks of monomials; the
constraints become $\alpha_1+\alpha_2=\ell_1$ and $|\alpha_2|+\alpha_3+2\alpha_4=\ell_2$, which
defines $\mathcal I_{\ell_1,\ell_2}$. For (GG), the monomials $u,u^2,uv,v,v^2$ give
$\alpha_1+2\alpha_2+\alpha_3=\ell_1$ and $\alpha_3+\alpha_4+2\alpha_5=\ell_2$, which defines
$\mathcal J_{\ell_1,\ell_2}$. For (HN), $d=0$ makes $\ell_1=0$ the unique empty index.
With $\ell_2=\alpha$, the equations are indexed exactly by $1\le\alpha\le r$, and
$\beta=(n_1,n_2)$ satisfies $n_1+2n_2=\alpha$. The shared (DS) $r_1$ system and the Gaussian-gated
system~(9) use this same scalar convention. For (SM) and (GG), the indices satisfy
$1\le|\ell_1|+\ell_2\le r$ (for (SM) this is written $0\le|\ell_1|\le r$,
$0\le\ell_2\le r-|\ell_1|$, $|\ell_1|+\ell_2\ge1$). The routed (DS) $r_2$ system uses this same
total-degree range specified in \cref{hn:def-moe}.
\end{proof}

\begin{lemma}[Rigidity of the extra variables]\label{hn:lem-rigidity}
In each item, suppose that the real parameters, with all weights $w_j>0$, solve the
corresponding order-$r$ system $\mathcal S_r$ of \cref{hn:def-moe}.
\begin{enumerate}[label=(\roman*)]
\item (HN) If $r\ge2$ and $b\equiv0$, then $a\equiv0$; if $r\ge4$ and $a\equiv0$, then $b\equiv0$.
\item (SM), any $d$: if $r\ge2$, then $p_1\equiv0$; if $r\ge4$, then also $p_2\equiv0$.
\item (GG): if $r\ge2\Mhn$, then $q_1\equiv q_2\equiv q_4\equiv q_5\equiv0$, and then $q_3\equiv0$.
\end{enumerate}
\end{lemma}

\begin{proof}
(i) $\hnE_2=\sum_jw_j(a_j^2/2+b_j)$, and if $a\equiv0$ then $\hnE_4=\sum_jw_jb_j^2/2$.
(ii) The coefficient of $u_i^2$ is $\sum_jw_jp_{1ji}^2/2$: only $\alpha_1=2e_i$ contributes, because
$\ell_2=0$ forces $\alpha_2=0$. Once $p_1\equiv0$, the coefficient of $u_i^2v^2$ is
$\sum_jw_jp_{2ji}^2/2$: only $\alpha_2=2e_i$, $\alpha_3=\alpha_4=0$ contributes.
(iii) Setting $u=0$ (resp.\ $v=0$) keeps exactly the equations with $\ell_1=0$ (resp.\ $\ell_2=0$),
and these form (HN) in $(p,q_4,q_5)$ (resp.\ in $(p,q_1,q_2)$) of order $r\ge2\Mhn$. Since all
$p_l\neq0$, the rigidity part of \cref{hn:thm-main} gives $q_4\equiv q_5\equiv0$ and
$q_1\equiv q_2\equiv0$. Then $F=\sum_lw_le^{q_{3l}uv}$, whose coefficient of $u^2v^2$ (total
degree $4\le2\Mhn$) is $\sum_lw_lq_{3l}^2/2$.
\end{proof}

\begin{corollary}[Orders of the mixture-of-experts systems]\label{hn:cor-moe}
Let $\Mhn\ge2$ and $d\ge1$, with all unknowns real and all weights non-zero.
Here $d$ is the model covariate dimension.
\begin{enumerate}[label=(\alph*)]
\item (SM) For $r\le2\Mhn-1$ there is a solution with all $p_{5j}\neq0$, some $p_{3j}\neq0$,
$p_4\not\equiv0$ and $p_1=p_2=0$. For $r\ge2\Mhn$ every solution with all $p_{5j}\neq0$ has
$p_1=p_2=p_3=p_4=0$. Hence $\rbar_{\mathrm{SM}}(\Mhn)=2\Mhn$ for every $d$, and the same value is
obtained under the weaker non-triviality condition $(p_1,p_2,p_3,p_4)\not\equiv0$.
\item (GG, $d=1$) For $r\le2\Mhn-1$ there is a solution with all $p_l\neq0$ and some
$q_{4l}\neq0$; for $r\ge2\Mhn$ every solution with all $p_l\neq0$ has $q\equiv0$. Hence
$\tilde r(\Mhn)=2\Mhn$; in the convention of \cref{hn:def-system}, the order of system~(9) of
the same paper is $\rbar(\Mhn-1)=2\Mhn$.
\item (DS) $r_1(\Mhn)=r_2(\Mhn)=2\Mhn$.
\end{enumerate}
\end{corollary}

\begin{proof}
\emph{Lower bound ($r\le2\Mhn-1$).} Let $(x_j,\pi_j)$ be the Gauss--Hermite rule. Give atom $j$ the weight $\pi_j$ (so
$p_{5j}=p_j=s_{3j}=t_{5j}=\sqrt{\pi_j}$), the coefficient $x_j$ to the monomial $v$, the
coefficient $-\frac12$ to $v^2$, and $0$ to all other monomials. Then
$F(u,v)=\sum_j\pi_je^{x_jv-v^2/2}$ does not depend on $u$, so every equation with $\ell_1\neq0$
holds, and the equations with $\ell_1=0$ form $\hnsys{\Mhn}{r}$, which holds for $r\le2\Mhn-1$ by
\cref{hn:prop-witness}. Since $\Mhn\ge2$, some $x_j\neq0$ (the coefficient of $v$, called $p_3$,
$q_4$, $s_1$, $t_3$ in the four systems), and the coefficient of $v^2$ (called $p_4$, $q_5$, $s_2$,
$t_4$) equals $-\frac12\neq0$. Thus the witness satisfies the non-triviality condition in each system.

\emph{Upper bound ($r\ge2\Mhn$).} (a) Let $r\ge2\Mhn$. Setting $u=0$ leaves (HN) in $(p_5,p_3,p_4)$ of order $r$,
so $p_3\equiv p_4\equiv0$ by \cref{hn:thm-main}, and \cref{hn:lem-rigidity}(ii) gives
$p_1\equiv p_2\equiv0$ because $2\Mhn\ge4$. (b) is \cref{hn:lem-rigidity}(iii) together with the
witness; system~(9) of \citet{nguyen2024gaussiangated} is (HN). (c) The $r_1$-system is (HN). For
the $r_2$-system, (a) shows that for $r\ge2\Mhn$ no solution with all $t_{5i}\neq0$ has some
$t_{4i}\neq0$, while the witness has $t_4\equiv-\frac12$.
\end{proof}

The restriction to $u=0$, which gives
$\rbar_{\mathrm{SM}}(\Mhn)\le\rbar(\Mhn-1)$ and
$\tilde r(\Mhn)\le\rbar(\Mhn-1)$, already appears in
\citet[proof of Lemma~1]{nguyen2023softmax} and in the proof of the corresponding lemma of
\citet{nguyen2024gaussiangated}; the new input is the value of $\rbar(\Mhn-1)$ and the extension
of the witness. The corollary also applies to the softmax-gated system used by
\citet{hai2025dendrogram}, who predict
$\rbar_{\mathrm{SM}}(\Mhn)=2\Mhn$ in fitted-expert-count indexing.

\begin{proposition}[Synchronized replication in two standard-softmax losses]
\label{hn:prop-softmax-replication}
Let $v=\tfrac14$, let the scalar covariate be uniform on $[-1,1]$, and write
$\hnphi_{\mu,s}$ for the Gaussian density with mean $\mu$ and variance $s>0$.
\begin{enumerate}[label=(\alph*),leftmargin=*]
\item For the standard-softmax Gaussian mixture-of-experts model, let the parameter space be
\[
\Theta_{\mathrm{SM}}
 =[-2,2]\times[-2,2]\times[-1,1]\times[-3,3]\times[1/2,3/2]
\]
in the coordinate order $(\beta_0,\beta_1,a,b,\sigma)$. The two true atoms are
\[
\omega_1^*=(0,-1,0,-2,1),\qquad
\omega_2^*=(0,1,0,2,1),
\]
and each $\omega_j^*$ is replaced by its unchanged copy and a copy with the same expert parameters
whose gate slope is increased by $v$. The resulting four-atom fit belongs to
$\mathcal O_4(\Theta_{\mathrm{SM}})$, satisfies the hypotheses of
\citet[Theorem~2]{nguyen2023softmax}, and has the two generating pairs as its
Voronoi cells, fixed before the translation infimum. Its conditional density equals the true
conditional density for every $(x,y)$, but the fixed-cell objective in
\citet[equation~(6)]{nguyen2023softmax} is
\[
F_{\mathrm{SM}}(t_1,t_2)
 =2\bigl(|t_2|^4+|v-t_2|^4\bigr)+2|2-e^{t_1}|.
\]
The parameter-preserving translation set is
\[
\begin{aligned}
T_{\mathrm{SM}}
 &=\bigl\{(t_1,t_2)\in\RR^2:
       (t_1,-1+t_2)\in[-2,2]^2,\ 
       (t_1,1+t_2)\in[-2,2]^2\bigr\}\\
 &=[-2,2]\times[-1,1],
\end{aligned}
\]
and the associated infima are
\[
\begin{aligned}
I_{\mathbb R}&:=\inf_{(t_1,t_2)\in\RR^2}F_{\mathrm{SM}}(t_1,t_2),\\
I_T&:=\inf_{(t_1,t_2)\in T_{\mathrm{SM}}}F_{\mathrm{SM}}(t_1,t_2).
\end{aligned}
\]
They are both attained at $(t_1,t_2)=(\log2,v/2)=(\log2,1/8)\in T_{\mathrm{SM}}$, and
\[
I_{\mathbb R}=I_T=\frac1{1024}>0.
\]
Thus, under either explicitly defined translation convention, the loss value is
\[
D_2^{\mathrm{SM}}=\frac1{1024}>0.
\]
\item In the dense standard-softmax model, one shared atom has probability weight one and
parameters $(\kappa_1,\kappa_0,\tau)=(0,0,1)$. The true routed atoms are
\[
\rho_1^*=(0,1,0,-2,1),\qquad
\rho_2^*=(0,0,0,2,1)
\]
in the coordinate order $(\beta_0,\beta_1,\eta_1,\eta_0,\nu)$, and each is replaced by its unchanged
copy and a same-expert copy whose gate slope is increased by $v$. The parameter spaces are
\[
\Theta_1=[1/2,3/2]\times[-1,1]\times[-1,1]\times[1/2,3/2],
\]
in the shared coordinate order
$(\text{weight},\kappa_1,\kappa_0,\tau)$, and
\[
\Theta_2=[-2,2]\times[-1,2]\times[-1,1]\times[-3,3]\times[1/2,3/2]
\]
in the routed coordinate order above, with fitted cardinality bounds $k_1=2$ and $k_2=4$.
The unchanged copy of $\rho_2^*$ is indexed last. The resulting configuration belongs to the fitted
class $\mathcal G_{2,4}(\Theta_1\times\Theta_2)$ and satisfies assumptions A.1--A.3 and the
linear-expert hypothesis of \citet[Theorem~2]{nguyen2025deepseek}. Its shared cell is unchanged,
and its routed cells are the two generating pairs. The complete conditional density equals the
true density for every $(x,y)$, whereas the loss of
\citet[equation~(6)]{nguyen2025deepseek} is
\[
D_2^{\mathrm{DS}}=\frac{257}{128}>0.
\]
\end{enumerate}
Consequently, coercive lower bounds in terms of the two $D_2$ losses fail over these fitted
classes; for the softmax loss this holds under both translation conventions. A parameter loss
that supports such a bound must identify synchronized replications or be used on a fitted class
that excludes them. A smaller local identifiable class is a separate setting for such a
loss comparison.
\end{proposition}

\begin{proof}
In \citet{nguyen2023softmax}, equations~(1), (2), and~(6) and Theorem~2 range over
$\mathcal O_k(\Theta)$; pairwise distinctness is imposed only on the true expert triples, so the
fitted class permits synchronized duplicate experts with distinct gate parameters. The coordinates
lie in $\Theta_{\mathrm{SM}}$, both variances are one, and the true expert triples
$(0,-2,1)$ and $(0,2,1)$ are distinct, with true gate slopes $-1$ and $1$. Together with the bounded
continuous covariate law, these properties verify the hypotheses, with $k_*=2<4=k$.

\citet{nguyen2023softmax} use two descriptions of the common translations. Their introduction describes arbitrary
$t_1\in\RR$ and $t_2\in\RR^d$ and relates the infimum in equations~(4) and~(6) to that invariance.
Section~3.1 explicitly restricts the $D_1$ infimum to translations keeping every translated true
gate parameter inside $\Theta$. Section~3.2 displays $D_2$ without restating the translation domain;
Theorem~2 adds no range, and the subsequent projected-gradient discussion does not specify the
projection set \citep[Introduction and Sections~3.1--3.2]{nguyen2023softmax}. We therefore evaluate both the unrestricted $\RR^2$ and the parameter-preserving
$T_{\mathrm{SM}}$ translation infima.

For each $j$, the unchanged fitted copy has distance zero from its generating true support point
in the metric of \citet[equation~(5)]{nguyen2023softmax}, and the shifted copy has distance $v$.
Every copy has distance at least $4$ from the other true center because the two expert intercepts
differ by $4$. Hence equation~(5) assigns exactly the two generating copies to each cell, before
the translation infimum is taken.

Put
\[
\phi_1(y\mid x)=\hnphi_{-2,1}(y),\qquad
\phi_2(y\mid x)=\hnphi_{2,1}(y).
\]
The true unnormalised numerator and denominator are
\[
N_*(x,y)=e^{-x}\phi_1(y\mid x)+e^x\phi_2(y\mid x),\qquad
Z_*(x)=e^{-x}+e^x.
\]
For the four fitted atoms they are
\[
\widehat N(x,y)=(1+e^{vx})N_*(x,y),\qquad
\widehat Z(x)=(1+e^{vx})Z_*(x).
\]
The common factor is positive, so $\widehat N/\widehat Z=N_*/Z_*$ pointwise and the expected
Hellinger distance is zero.

The only external order value needed to evaluate this loss is
$\rbar_{\mathrm{SM}}(2)=4$; \citet[Lemma~1]{nguyen2023softmax} states
$\rbar_{\mathrm{SM}}(2)=4$ and $\rbar_{\mathrm{SM}}(3)=6$ in every positive covariate dimension. Each Voronoi cell contains two fitted atoms. Their fitted gate-intercept weights are both one, their gate-slope offsets are
$0$ and $v$, and every expert difference vanishes. The mass discrepancy in each cell is
$|2-e^{t_1}|$. Substitution in equation~(6) therefore gives the objective $F_{\mathrm{SM}}$ in part~(a).

Both true gate intercepts are zero, so preserving their membership in $[-2,2]$ requires
$t_1\in[-2,2]$. For the true slopes $-1$ and $1$, the respective conditions are
$t_2\in[-1,3]$ and $t_2\in[-3,1]$. Their intersection is $[-1,1]$, which gives the formula
for $T_{\mathrm{SM}}$.

The last term of $F_{\mathrm{SM}}$ is nonnegative and vanishes at $t_1=\log2$. For every $t\in\RR$,
\[
t^4+(v-t)^4
 \ge\frac{\bigl(t^2+(v-t)^2\bigr)^2}{2}
 \ge\frac{v^4}{8},
\]
and both inequalities are equalities at $t=v/2$. Thus
\[
I_{\mathbb R}=\frac{v^4}{4}=\frac1{1024}.
\]
The minimizing pair is $(\log2,v/2)=(\log2,1/8)$. Since $0<\log2<1<2$ and
$1/8\in[-1,1]$, this pair belongs to $T_{\mathrm{SM}}$. As $T_{\mathrm{SM}}\subset\RR^2$,
\[
I_{\mathbb R}\le I_T
 \le F_{\mathrm{SM}}(\log2,1/8)=I_{\mathbb R}.
\]
Both infima are therefore attained at that pair and have the same positive value. The pointwise
density identity gives the same lower-bound obstruction under either convention.

For the dense standard-softmax model of Section~3, A.3 separates only the true experts, and
A.2 fixes only the true reference logit. The fitted class allows coincident experts;
Appendix~C.2 additionally assumes a fitted reference normalization that is absent from the class
and Theorem~2 \citep{nguyen2025deepseek}. In our configuration the shared atom is fixed with the
required probability weight, all routed masses are $e^{\beta_0}=1$, and all coordinates lie in
their respective parameter boxes. The true routed experts $(\eta_1,\eta_0)=(0,-2),(0,2)$ are
distinct and all variances are one. Indexing the unchanged second routed atom last makes both
the true and fitted last routed gates $(\beta_0,\beta_1)=(0,0)$, while the other true slope is
nonzero. Since the covariate interval is bounded, the linear experts are bounded and Lipschitz.
Together these properties verify A.1--A.3 and the linear-expert condition.

The shifted copy in each routed cell is at distance $v$ from its generating center and at least
$4$ from the other center because the expert intercepts differ by $4$. Hence the two routed
Voronoi cells are precisely the generating pairs, while the shared cell contains its unchanged
atom. The true routed numerator and denominator are
\[
N_{\mathrm r,*}(x,y)=e^x\phi_1(y\mid x)+\phi_2(y\mid x),\qquad
Z_{\mathrm r,*}(x)=e^x+1.
\]
The fitted routed quantities satisfy
\[
\widehat N_{\mathrm r}(x,y)=(1+e^{vx})N_{\mathrm r,*}(x,y),\qquad
\widehat Z_{\mathrm r}(x)=(1+e^{vx})Z_{\mathrm r,*}(x).
\]
Thus the routed conditional density is unchanged. The shared density is unchanged separately, so
the complete half-shared plus half-routed density is identical pointwise and its expected total
variation distance is zero.

The only external order value needed here is $r_2(2)=4$
\citep[Appendix~A, Proposition~4]{nguyen2025deepseek}.
Every shared term in equation~(6) vanishes. In each routed cell the mass
discrepancy is $|1+1-1|=1$. The unchanged copy has zero parameter error, while the shifted copy has
only the gate-slope difference $v$, contributing $v^4$. Therefore
\[
D_2^{\mathrm{DS}}=2(1+v^4)=2+\frac2{256}=\frac{257}{128}.
\]
The claimed lower bounds would have zero left-hand side and a strictly positive right-hand side on
these examples, which proves the final assertions.
\end{proof}

\begin{remark}[Candidate rates and regularity issues for Gaussian-gated mixtures and related models]\label{hn:rem-moe-rates}
Theorems~1 and~2 of \citet{nguyen2024gaussiangated} report parameter-rate expressions.
Formally inserting the independently proved coefficient order $2\Mhn$ gives, in the
Type~I case $c_j^0\neq0$ for every true gate, the candidate exponents
$n^{-1/(4\Mhn)}$ for $c_j^0,b_j^0$, $n^{-1/(2\Mhn)}$ for
$\Gamma_j^0,\nu_j^0$, and $n^{-1/4}$ for $a_j^0$. In the Type~II setting,
$d=1$ and the relevant cells have $c_j^0=0$; the corresponding candidate expressions are
$n^{-1/(4\Mhn)}$ for $c,b$ and $n^{-1/(2\Mhn)}$ for $\Gamma,a,\nu$.
Turning these candidate exponents into statistical convergence bounds requires resolving
three issues in the arguments of \citet{nguyen2024gaussiangated}. For the derivative family
in equation~(18), write $f_L(x)=\hnphi_{c,\Gamma}(x)$ and
$f_D(y;h)=\hnphi_{h,\nu}(y)$ with $h=ax+b$ and $\Gamma,\nu>0$.
In its scalar specialization, the allowed indices satisfy
$0\le\alpha\le q$ and $\tau_1+\tau_2\le2(q-\alpha)$, where $q\ge1$.
The indices $(\alpha,\tau_1,\tau_2)=(1,0,0),(0,0,1),(0,1,1)$ therefore give the three members
$x f_L\partial_h f_D$, $f_L\partial_h f_D$, and $(\partial_c f_L)\partial_h f_D$.
Since $\partial_c f_L=(x-c)f_L/\Gamma$, they satisfy the exact identity
\[
 (x f_L-c f_L-\Gamma\partial_c f_L)\partial_h f_D=0.
\]
Thus the claimed linear independence of that indexed family in Appendix~B.1 fails already
in dimension one, including the Type~I case $c\ne0$. A fitted-weight floor does not remove
this identity. Separately, the expert-index separations after equations~(29) and~(43) in
Appendices~B.1 and~B.2 assume pairwise-distinct true expert triples $(a_j^0,b_j^0,\nu_j^0)$,
which are not implied by distinct full atoms. Adding this missing assumption does not remove
the derivative identity.

Finally, the ordinary nondegenerate Gaussian law for $X$ in equation~(2) cannot also have the
bounded support used in the density-rate proof in Appendix~C.2. On a bounded Borel set
$\mathcal C$, the restricted Gaussian has mass
$Z_{\mathcal C}(c,\Gamma)=\int_{\mathcal C}f_L(x)\,\dd x<1$.
When this mass is positive, normalizing the restriction changes the family and, wherever the
parameter derivative exists, gives on $\mathcal C$
\[
 \partial_c\widetilde f_{L,\mathcal C}
 =\left(\frac{x-c}{\Gamma}-\partial_c\log Z_{\mathcal C}(c,\Gamma)\right)
   \widetilde f_{L,\mathcal C},\qquad
 \widetilde f_{L,\mathcal C}=f_L/Z_{\mathcal C}(c,\Gamma).
\]
A truncated model therefore requires its own Taylor expansion and coercive loss comparison,
including the normalizer derivatives.

The cited versions also contain two scaling inconsistencies. In
\citet[Appendix~A.2, immediately after equation~(29)]{nguyen2023softmax}, $\overline M_n$ is defined
using $\|\Delta a\|^{1/2}$ and $|\Delta\sigma|^{1/2}$, but the corresponding limits divide
$\Delta a$ and $\Delta\sigma$ by $\overline M_n$ and $2\overline M_n$, respectively. Sections~3.2 and~4.2 of \citet{nguyen2025deepseek} state sample complexity
$O(\epsilon^{-r_2(m)})$ but specialize it to $O(\epsilon^{-12})$ when $m=3$, although Appendix~A
gives $r_2(3)=6$; the preceding full-expert rate $n^{-1/(2r_2)}$ instead yields
$O(\epsilon^{-2r_2})$, which is consistent with the numerical exponent.
\end{remark}

\hnsection{Statistical consequences}\label{hn:sec-stat}

The notation below follows \citet{ho2015arxiv,ho2016aos}. $\Okc$ is the set of mixing measures on
$\Theta\times\Omega$ with at most $k$ atoms, each of weight at least $c_0$, and $\mathcal E_{k_0}$
is the set of those with exactly $k_0$ atoms. $\Wass r$ denotes the Wasserstein distance between
mixing measures with ground cost $\|\theta-\theta'\|+\|\Sigma-\Sigma'\|$ (Euclidean and entrywise
$\ell_2$ norms).

\begin{fact}[{\citealp[Theorem~1.1(b)]{ho2016aos}}]\label{hn:fact-hn-mle}
Let $L,\gamma>0$ and $0<\underline\lambda<\overline\lambda$ be fixed, $\Theta=[-a_n,a_n]^d$ with
$a_n\le L(\log n)^\gamma$, and $\Omega\subset\mathbb S^d_{++}$ compact with all eigenvalues in
$[\underline\lambda,\overline\lambda]$. Let $c_0=0$ if $k-k_0\in\{1,2\}$ and $0<c_0<1$ otherwise.
Assume $G_0\in\mathcal E_{k_0}\cap\Okc$, and let $\Ghat$ be the maximum-likelihood estimator over
$\Okc$. Then
$\PP\big(\Wass{\rbar}(\Ghat,G_0)>C(\log n/n)^{1/(2\rbar)}\big)\lesssim\exp(-c\log n)$, with
$\rbar=\rbar(k-k_0)$ and constants $C,c>0$ depending only on $d,L,\gamma,\underline\lambda,
\overline\lambda,c_0$ and $G_0$.
\end{fact}

\begin{corollary}[Pointwise rate of the maximum-likelihood estimator]\label{hn:cor-mle}
Under the hypotheses of \cref{hn:fact-hn-mle}, with $\Mhn=k-k_0+1\ge2$,
\[
\PP\Big(\Wass{2\Mhn}(\Ghat,G_0)>C\Big(\frac{\log n}{n}\Big)^{1/(4\Mhn)}\Big)\lesssim\exp(-c\log n);
\]
in particular
$\Wass1(\Ghat,G_0)\le\Wass{2\Mhn}(\Ghat,G_0)=O_\PP\big((\log n/n)^{1/(4\Mhn)}\big)$.
The constants depend on $G_0$: the rate is pointwise, not uniform over $\Okc$.
\end{corollary}

\begin{proof}
Insert $\rbar(k-k_0)=2\Mhn$ (\cref{hn:thm-main}) into \cref{hn:fact-hn-mle}, and use
$\Wass1\le\Wass r$ for probability measures.
\end{proof}

Returning to the setting of \citet{ho2015arxiv}, let
$\Theta$ be a fixed bounded location set, let $\Omega$ be a positive-definite covariance space
whose covariance eigenvalues are uniformly bounded above, assume that the same fixed true mixing
measure satisfies $G_0\in\mathcal E_{k_0}(\Theta\times\Omega)$, and fix
$\underline c\in(0,1)$. Write $\mathcal O_{k,\underline c}$ for the fitted mixing measures
with at most $k$ atoms and every fitted weight at least $\underline c$. In this setting,
\citet[Theorem~4.1(b)]{ho2015arxiv} with $\rbar=2\Mhn$ gives the local inverse bound
\[
\TV(p_G,p_{G_0})\gtrsim\Wass{2\Mhn}^{2\Mhn}(G,G_0)
\]
for $G\in\mathcal O_{k,\underline c}$ near the fixed $G_0$. Separately,
\citet[Theorem~4.1(a)]{ho2015arxiv} states a floor-free Hellinger sharpness infimum over
$\mathcal O_k$. When the perturbed parameters remain in $\Theta\times\Omega$---for instance, when the split
atom is interior---the Gauss--Hermite path in Proposition~\ref{hn:prop-admissible-sharpness}
certifies sharpness of the exponent $2\Mhn$. The local floor-free extension of \citet[Proposition~4.3]{ho2015arxiv} applies only
when $k-k_0$ is one or two. Locations and scales receive the same rate in $\Wass{2\Mhn}$;
separate rates require a finer loss (\cref{hn:rem-variance}).

\begin{proposition}[Sharpness along an admissible atom-splitting path]
\label{hn:prop-admissible-sharpness}
Let $d\ge1$, $k_0\ge1$, and $\Mhn\ge2$ be integers, and put $k=k_0+\Mhn-1$. Let
$\Theta\subset\RR^d$ and $\Omega\subset\mathbb S^d_{++}$, and let
\[
 G_0=\sum_{i=1}^{k_0}p_i^0\delta_{\omega_i^0},\qquad
 \omega_i^0=(\theta_i^0,\Sigma_i^0)\in\Theta\times\Omega,
 \qquad p_i^0>0,\quad\sum_i p_i^0=1,
\]
have distinct atoms. Use the ground metric
\[
 d_{\rm par}((\theta,\Sigma),(\theta',\Sigma'))
   =\|\theta-\theta'\|_2+\|\Sigma-\Sigma'\|_{\rm F}.
\]
Let $(x_j,\pi_j)_{j=1}^{\Mhn}$ be the nodes and probability weights of the probabilists'
Gauss--Hermite rule, put
$R=(\Sigma_1^0)^{1/2}$ and $r=Re_1$, where $e_1$ is the first standard basis vector, and define
\[
 \gamma_j(t)=\bigl(\theta_1^0+tx_jr,\ \Sigma_1^0-t^2rr^\top\bigr).
\]
Assume that
\[
 \tag{P}\label{hn:eq-admissible-path}
 \gamma_j(t)\in\Theta\times\Omega
 \quad(1\le j\le\Mhn,\ 0\le t\le t_{\rm path})
 \qquad\text{for some }0<t_{\rm path}<1.
\]
For $0\le t\le t_{\rm path}$ put
\[
 G_t=\sum_{i=2}^{k_0}p_i^0\delta_{\omega_i^0}
       +p_1^0\sum_{j=1}^{\Mhn}\pi_j\delta_{\gamma_j(t)},
\]
where the first sum is absent for $k_0=1$ and coincident descendants are merged at $t=0$.
Write $\mathcal O_k(\Theta\times\Omega)$ for probability mixing measures with at most $k$ atoms
in this product, and $\mathcal O_{k,c}(\Theta\times\Omega)$ for those whose every support weight
is at least $c$. Set
\[
 c_{\rm path}=
 \min\bigl(\{p_i^0:2\le i\le k_0\}
       \cup\{p_1^0\pi_j:1\le j\le\Mhn\}\bigr)>0,
\]
omitting the first set when $k_0=1$. There are $t_0,c_W,C_W>0$, with
$t_0\le\min\{t_{\rm path},1/2\}$, such that for $0<t\le t_0$ the measure $G_t$ has exactly
$k$ atoms and
\[
 0<c_Wt\le\Wass1(G_t,G_0)\le C_Wt,\qquad
 \chisq(p_{G_t}\|p_{G_0})\le p_1^0C_\Mhn t^{4\Mhn},\qquad
 \Hel(p_{G_t},p_{G_0})\le\sqrt{p_1^0C_\Mhn/2}\,t^{2\Mhn}.
\]
Here $C_\Mhn$ is the constant in \cref{hn:lem-chi2}, and
$\Hel^2(p,q)=\frac12\int_{\RR^d}(\sqrt p-\sqrt q)^2$.
Consequently, for every real $1\le q<2\Mhn$ and every $\epsilon>0$,
\[
 \inf_{\substack{G\in\mathcal O_k(\Theta\times\Omega)\\
                 0<\Wass1(G,G_0)\le\epsilon}}
 \frac{\Hel(p_G,p_{G_0})}{\Wass1(G,G_0)^q}=0.
\]
The same equality holds over $\mathcal O_{k,c}(\Theta\times\Omega)$ for every fixed
$0<c\le c_{\rm path}$. Condition
\eqref{hn:eq-admissible-path} follows if the single split atom $\omega_1^0$ is an interior point
of $\Theta\times\Omega$ in $\RR^d\times\mathbb S^d$. The other true atoms need not be interior.
\end{proposition}

\begin{proof}
The covariance along the path is
\[
 \Sigma_1^0-t^2rr^\top=R(I_d-t^2e_1e_1^\top)R.
\]
It is positive definite for $0\le t<1$, since $R$ is invertible and the middle matrix has
eigenvalues $1-t^2,1,\ldots,1$. Membership in the possibly smaller set $\Omega$, and membership of
the means in $\Theta$, use \eqref{hn:eq-admissible-path}. Distinct Gauss--Hermite nodes and
$r\ne0$ make the descendants distinct for $t>0$. Each descendant tends to $\omega_1^0$.
Choose $\eta>0$ smaller than the distance from $\omega_1^0$ to every unchanged true atom,
with $\eta=1$ if there are no unchanged atoms. After decreasing $t_0$, every descendant has
distance less than $\eta$ from $\omega_1^0$ and so differs from every unchanged atom. Hence
$G_t$ has exactly $(k_0-1)+\Mhn=k$ atoms. Their weights are $p_i^0$ for $i\ge2$ and
$p_1^0\pi_j$; their minimum is $c_{\rm path}$. At $t=0$ they merge to $p_1^0$, which is no smaller
than any $p_1^0\pi_j$. Thus, for all sufficiently small $t\ge0$ and every $0<c\le c_{\rm path}$,
$G_t\in\mathcal O_{k,c}(\Theta\times\Omega)$.

The direct coupling sends each descendant to $\omega_1^0$ and leaves all other atoms fixed.
For $0<t\le1$ it gives
\[
 \Wass1(G_t,G_0)
 \le p_1^0\sum_j\pi_j
       \bigl(t|x_j|\|r\|_2+t^2\|rr^\top\|_{\rm F}\bigr)
 \le C_Wt,
\]
where $C_W=p_1^0\sum_j\pi_j(|x_j|\|r\|_2+\|rr^\top\|_{\rm F})$.
For the lower bound, the function
\[
 \psi(\omega)=\min\{d_{\rm par}(\omega,\omega_1^0),\eta\}
\]
is $1$-Lipschitz. Integrating its difference against any coupling and then taking the infimum
over couplings gives
\[
 \Wass1(G_t,G_0)\ge
 \left|\int\psi\,\dd G_t-\int\psi\,\dd G_0\right|.
\]
The unchanged atoms cancel in this difference. On each descendant the distance is below $\eta$,
so
\[
 \Wass1(G_t,G_0)
 \ge p_1^0\sum_j\pi_jd_{\rm par}(\gamma_j(t),\omega_1^0)
 \ge p_1^0\|r\|_2\Bigl(\sum_j\pi_j|x_j|\Bigr)t=:c_Wt.
\]
Quadrature exactness gives $\sum_j\pi_jx_j^2=1$, and therefore $c_W>0$.

Let $q_0$ be the density of $\Normal(\theta_1^0,\Sigma_1^0)$ and let $q_t$ be the normalized
descendant-cluster density. The unchanged mixture terms imply
\[
 p_{G_t}-p_{G_0}=p_1^0(q_t-q_0),\qquad p_{G_0}\ge p_1^0q_0.
\]
It follows directly from the defining integral that
\[
 \chisq(p_{G_t}\|p_{G_0})\le p_1^0\chisq(q_t\|q_0).
\]
The invertible change of variables $y=R^{-1}(x-\theta_1^0)$ leaves chi-square divergence
unchanged. It transforms $q_0$ into $\Normal(0,I_d)$ and $q_t$ into
\[
 \left(\sum_j\pi_j\Normal(tx_j,1-t^2)\right)\otimes\Normal(0,I_{d-1}),
\]
with the last factor omitted for $d=1$. A common tensor probability factor cancels in the
chi-square integral, so \cref{hn:lem-chi2} gives the stated $p_1^0C_\Mhn t^{4\Mhn}$ bound.
For positive densities,
\[
 (\sqrt p-\sqrt q)^2
   =\frac{(p-q)^2}{(\sqrt p+\sqrt q)^2}\le\frac{(p-q)^2}{q}.
\]
Thus $\Hel^2(p,q)\le\chisq(p\|q)/2$, which proves the Hellinger bound.

For $1\le q<2\Mhn$, the two path estimates give
\[
 \frac{\Hel(p_{G_t},p_{G_0})}{\Wass1(G_t,G_0)^q}
 \le\sqrt{p_1^0C_\Mhn/2}\,c_W^{-q}t^{2\Mhn-q}\longrightarrow0.
\]
For every sufficiently small $t>0$, $\Wass1(G_t,G_0)$ is strictly positive and tends to zero as
$t\downarrow0$, so the path enters every prescribed positive neighborhood. The quotients are nonnegative and
the same path belongs to each floor class with $c\le c_{\rm path}$, establishing both infimum
statements.

Finally, if $\omega_1^0$ is interior in the ambient product space, an open ball about it lies in
$\Theta\times\Omega$. Continuity of the finitely many curves $\gamma_j$ supplies one common
positive interval on which they all lie in that ball. Decreasing its endpoint below $1$ proves
\eqref{hn:eq-admissible-path}. All other true atoms stay fixed, so no interiority assumption on
them is used.
\end{proof}
\begin{proposition}[Exact local minimax order on a shrinking neighbourhood]
\label{hn:prop-localminimax}
Let $d\ge1$, $k_0\ge1$, $\Mhn\ge2$, $k=k_0+\Mhn-1$, and
$\beta=1/(4\Mhn)$. Let
$\Theta\times\Omega\subset\RR^d\times\mathbb S^d_{++}$ be compact, and let
\[
G_0=\sum_{i=1}^{k_0}p_i^0\delta_{(\theta_i^0,\Sigma_i^0)}
 \in\mathcal E_{k_0}
\]
have distinct atoms in the interior of $\Theta\times\Omega$. Put
\[
c_0=\left(\min_i p_i^0\right)\left(\min_j\pi_j\right)>0,
\]
where $(\pi_j)$ are the weights of the $\Mhn$-point Gauss--Hermite rule. There are
$C_1,c,n_0>0$, depending on $G_0$, $\Mhn$, and $\Theta\times\Omega$, such that for every
$n\ge n_0$,
\[
c\,n^{-\beta}
\le
\inf_{\widehat G}\ \sup_{G\in\mathcal N_n}
 \EE_G\,\Wass1(\widehat G,G)
\le C_1n^{-\beta},
\qquad
\mathcal N_n=
 \big\{G\in\Okc:\Wass1(G,G_0)\le C_1n^{-\beta}\big\}.
\]
Here $\EE_G$ refers to $X_1,\dots,X_n$ i.i.d.\ from $p_G$, and the infimum is over all measurable
estimators from $(\RR^d)^n$ to the probability measures on $\Theta\times\Omega$ equipped with
$\Wass1$. The upper bound uses the constant estimator at the fixed known center $G_0$; it is not
a uniform full-class estimation theorem.
\end{proposition}

\begin{proof}
Let $G_t$ be the path in Proposition~\ref{hn:prop-admissible-sharpness}, with its ground metric
$d_{\rm par}$, the Euclidean/Frobenius metric used for $\Wass1$ in this section. Since the atoms
of $G_0$ are interior to $\Theta\times\Omega$, condition~\eqref{hn:eq-admissible-path} holds for
some $t_{\rm path}\in(0,1)$. Since $\min_j\pi_j\le1$, the definitions give
$c_0\le c_{\rm path}$, so $G_0,G_t\in\Okc$ for all sufficiently small $t\ge0$. By Proposition~\ref{hn:prop-admissible-sharpness}, there are $t_*\in(0,1/2]$ and
$c_W,C_W>0$ such that, for $0<t\le t_*$,
\[
c_Wt\le\Wass1(G_t,G_0)\le C_Wt,\qquad
\chisq(p_{G_t}\|p_{G_0})\le p_1^0C_\Mhn t^{4\Mhn},
\]
where $C_\Mhn$ is the constant of \cref{hn:lem-chi2}.

Now let $\widehat G$ be an estimator, $\delta=\Wass1(G_0,G_t)/2$ and
$A=\{\Wass1(\widehat G,G_0)\ge\delta\}$. On $A^c$, $\Wass1(\widehat G,G_t)>\delta$ by the triangle
inequality. Writing $P^n_t$ for the law of the sample under $G_t$,
$\EE_{G_0}\Wass1(\widehat G,G_0)+\EE_{G_t}\Wass1(\widehat G,G_t)\ge\delta\{P_0^n(A)+P_t^n(A^c)\}
\ge\delta\{1-\TV(P_0^n,P_t^n)\}$. By Cauchy--Schwarz, $\TV(P,Q)\le\frac12\chisq(P\|Q)^{1/2}$, and
$1+\chisq(P^{\otimes n}\|Q^{\otimes n})=(1+\chisq(P\|Q))^n$. Put
\[
A_\Mhn=\left(\frac{\log2}{p_1^0C_\Mhn}\right)^{1/(4\Mhn)},
\qquad
t_n=A_\Mhn n^{-1/(4\Mhn)}.
\]
Then
$n\chisq(p_{G_{t_n}}\|p_{G_0})\le\log2$, so $\chisq(P_{t_n}^n\|P_0^n)\le e^{\log2}-1=1$ and
$\TV(P_0^n,P_{t_n}^n)\le\frac12$. Therefore
$\max_{G\in\{G_0,G_{t_n}\}}\EE_G\Wass1(\widehat G,G)\ge\frac\delta4\ge\frac{c_W}8t_n$. For $n\ge n_0$
we have $t_n\le\min\{t_*,\frac12\}$. Set
\[
C_1=C_WA_\Mhn,\qquad c=\frac{c_WA_\Mhn}{8}.
\]
The upper transport bound gives
$\Wass1(G_{t_n},G_0)\le C_1n^{-1/(4\Mhn)}$, so both $G_0$ and $G_{t_n}$ belong to
$\mathcal N_n$. Taking the infimum over estimators in the preceding two-point bound proves the
lower inequality.

For the upper inequality, use the constant estimator
\[
\widehat G_0(X_1,\dots,X_n)=G_0.
\]
It is measurable, and for every $G\in\mathcal N_n$,
\[
\EE_G\Wass1(\widehat G_0,G)
 =\Wass1(G_0,G)\le C_1n^{-1/(4\Mhn)}.
\]
Taking the supremum over $\mathcal N_n$ and then the infimum over all estimators proves the
matching upper bound.
\end{proof}

The minimax conclusion applies to shrinking neighbourhoods of the fixed known center $G_0$.
The uniform problem over $\Okc$ remains open \citep[Section~3.4]{nguyen2026ot}. The lower bound also reaches the
endpoint excluded by the earlier general lower bound. With $\rbar=2\Mhn$,
\citet[Theorem~1.1(a)]{ho2016aos} give the lower bound
$c\,n^{-1/r}$ for every $r<4\Mhn$, and
\citet[Proposition~9]{wu2020dmm} proved the rate $n^{-1/(4\Mhn)}$ in the
homoscedastic submodel with unknown common variance by the same Gauss--Hermite construction.
Together with Corollary~\ref{hn:cor-mle}, the lower bound shows that the exponent
$1/(4\Mhn)$ of the pointwise maximum-likelihood rate in $\Wass1$ cannot be improved uniformly
over these neighbourhoods, up to the logarithmic factor.

\begin{remark}[Separate rates for locations and scales]\label{hn:rem-variance}
\Cref{hn:cor-mle} gives one rate for locations and scales jointly. A split is available for a
penalised maximum-likelihood estimator. Manole--Ho index their order by the cell size. For
$l\ge2$, write $\bar r_{\mathrm{MH}}(l)=\rbar(l-1)$ in our redundant-component convention, so
that the order formula gives $\bar r_{\mathrm{MH}}(l)=2l$.

For this result, put $\mathcal P=[-a,a]^d\times\Omega$, a compact parameter domain with nonempty
interior whose covariance eigenvalues lie in a fixed closed subinterval of $(0,\infty)$. Let
$\widehat G_n^{\mathrm{pen}}$ denote the Manole--Ho penalised estimator
\[
\widehat G_n^{\mathrm{pen}}\in
\operatorname*{arg\,max}_{G\in\mathcal O_k(\mathcal P)}
\left\{\ell_n(G)+(\log n)\rho(G)\right\},
\qquad
\ell_n(G)=\sum_{m=1}^n\log p_G(X_m),
\qquad
\rho(G)=\sum_{i=1}^{|\supp G|}\log p_i .
\]
Here $\mathcal O_k(\mathcal P)$ is the class of probability measures on $\mathcal P$ with at most
$k$ distinct support atoms, and the $p_i$ in $\rho(G)$ are their strictly positive weights.
Write the distinct support atoms of $\widehat G_n^{\mathrm{pen}}$ as
$(\widehat\mu_i,\widehat\Sigma_i)$, and let $\mathcal A_j$ be their Voronoi cells relative to the
atoms of the fixed true mixing measure. \citet[Theorem~5 and its consequence~(i)]{manole2022refined}
show that
$\|\widehat\mu_i-\mu_j^0\|\lesssim(\log n/\sqrt n)^{1/\bar r_{\mathrm{MH}}(|\mathcal A_j|)}$ and
$\|\widehat\Sigma_i-\Sigma_j^0\|\lesssim
(\log n/\sqrt n)^{2/\bar r_{\mathrm{MH}}(|\mathcal A_j|)}$ with probability tending to one, for
atoms $i$ in a cell $\mathcal A_j$ with $|\mathcal A_j|\ge2$. Thus the exponents of
$\log n/\sqrt n$ are $1/(2|\mathcal A_j|)$ for locations and $1/|\mathcal A_j|$ for scales.
These are pointwise upper bounds under their conditions; the constants may depend on the fixed
true mixing measure and the model parameters. No separate scale rate follows from the
Wasserstein conclusions in \citet[Theorem~4.1(b)]{ho2015arxiv} or
\citet[Theorem~1.1(b)]{ho2016aos}. For univariate location--scale Gaussian o-mixtures over the
fitted-weight-floored class,
\citet[Proposition~3.3]{ho2019simods} give singularity index
$(\rbar,\rbar/2)$ when $\rbar$ is even and
$(\rbar,(\rbar+1)/2)$ when $\rbar$ is odd; their general framework converts a root-$n$
Hellinger density rate into the corresponding coordinate rates.
\end{remark}

\hnsection{Remarks on the hypotheses and the literature}\label{hn:sec-remarks}

\begin{remark}[Complex, signed-weight, and zero-weight variants]\label{hn:rem-hypotheses}
(1) \emph{Jointly complex coefficient unknowns.} Allow $(a_j,b_j,c_j)$ to lie in $\CC$, and let
\[
\omega=e^{2\pi i/3},\qquad
z=e^{\pi i/3}=-\omega^2,\qquad
q=\frac{\omega-\omega^2}{6}=\frac{i}{2\sqrt3}.
\]
For $\Mhn=2$, take
\[
c=(1,z),\qquad w=(c_1^2,c_2^2)=(1,\omega),\qquad
a=(1,z),\qquad
b=(q,-\omega q)
 =\left(\frac{i}{2\sqrt3},\frac14+\frac{i}{4\sqrt3}\right).
\]
The relations
\[
\omega^2+\omega+1=0,\qquad \omega^3=1,\qquad
z^2=\omega,\qquad z^3=-1,\qquad z^4=\omega^2,\qquad
q^2=-\frac1{12}
\]
follow directly from the definitions. We shall also use
\[
(\omega-\omega^2)(1-\omega^2)=3\omega,
\qquad
q(1-\omega^2)=\frac{\omega}{2},
\qquad
-\omega^2q=\frac{\omega-1}{6}.
\]
Expanding $\exp(a_jt+b_jt^2)$ as a formal power series in $t$ gives
\[
\begin{aligned}
\hnE_1&=\sum_jw_ja_j,\\
\hnE_2&=\sum_jw_j\left(b_j+\frac{a_j^2}{2}\right),\\
\hnE_3&=\sum_jw_j\left(a_jb_j+\frac{a_j^3}{6}\right),\\
\hnE_4&=\sum_jw_j\left(\frac{b_j^2}{2}
 +\frac{a_j^2b_j}{2}+\frac{a_j^4}{24}\right).
\end{aligned}
\]
The four coefficients therefore vanish by the exact calculations
\[
\begin{aligned}
\hnE_1
 &=1+\omega z=1-\omega^3=0,\\
\hnE_2
 &=q+\frac12+\omega\left(-\omega q+\frac{\omega}{2}\right)
  =q(1-\omega^2)+\frac{1+\omega^2}{2}
  =\frac{\omega}{2}-\frac{\omega}{2}=0,\\
\hnE_3
 &=q+\frac16+\omega\left(z(-\omega q)+\frac{z^3}{6}\right)
  =(1+\omega)q+\frac{1-\omega}{6}
  =-\omega^2q+\frac{1-\omega}{6}=0,\\
\hnE_4
 &=\left(\frac{q^2}{2}+\frac q2+\frac1{24}\right)
  +\omega\left(\frac{\omega^2q^2}{2}
    -\frac{\omega^2q}{2}+\frac{\omega^2}{24}\right)\\
 &=\left(\frac{q^2}{2}+\frac q2+\frac1{24}\right)
  +\left(\frac{q^2}{2}-\frac q2+\frac1{24}\right)
  =q^2+\frac1{12}=0.
\end{aligned}
\]
Here $c_1=1\neq0$, $c_2=z\neq0$ because $z^3=-1$, and $a_1=1\neq0$.
Thus the real rigidity conclusion of \cref{hn:thm-main} fails over $\CC$ already for
$\Mhn=2$. More trivially, $c_2=ic_1$ with two
identical atoms solves every order. (2) \emph{Signed weights.} If the weights $w_j$ were real but
allowed to be negative, $w=(1,-1)$ with two identical atoms $(a_1,b_1)=(a_2,b_2)\neq0$ would solve
every order; positivity of $w_j=c_j^2$ is what excludes this cancellation in
\cref{hn:prop-moment}. (3) \emph{Zero weights.} If some $c_j=0$ is allowed, solutions with
$a\neq0$ exist at every order
\citep[remark following system~(8)]{ho2015arxiv}.
\end{remark}

\begin{remark}[Explicit witnesses and normalizations in the literature]\label{hn:rem-errata}
(i) The order-$5$ witness given for $\Mhn=3$ in
\citet[proof of Lemma~1]{nguyen2023softmax}, namely
$(p_3,p_4)=(\pm\frac{\sqrt3}3,-\frac16)$ and $(0,0)$ with unit weights, violates the
$(\ell_1,\ell_2)=(0,4)$ equation, whose value is
$2\big(\frac1{216}-\frac1{36}+\frac1{72}\big)=-\frac1{54}$. The Gaussian-gated formulation of \citet{nguyen2024gaussiangated} has separate
discrepancies: its zero-weight tuple has $p_l=0$, contrary to its non-triviality
condition, and its $(0,4)$ equation has $q_{4l}^2/4!$ in place of the $q_{4l}^4/4!$
required by system~(12). The values
$\rbar_{\mathrm{SM}}(3)=\tilde r(3)=6$ are nevertheless proved here: the three-point Gauss--Hermite witness
solves the system to order $5$, with first non-zero coefficient
$[u^0v^6]F=\hnE_6=-\frac{3!}{6!}=-\frac1{120}$.

(ii) In the proof of Proposition~4.2 of \citet{ho2015arxiv}, the normalisations are
$b=b_1/a_1$ and $c=b_2/a_1$. Dimensional consistency with the weighted
scaling used in that proof requires division by $a_1^2$. The proof of \cref{hn:thm-main} is independent of this normalization.
\end{remark}

\begin{remark}[Notes on cited versions]\label{hn:rem-numbering}
Numbered citations refer to the versions listed in the bibliography. The coefficient system
and local inverse comparison use Technical Report~536 of \citet{ho2015arxiv},
arXiv:1501.02497v1. The references to \citet{ho2016aos,ho2016ejs} use the linked author
manuscripts. For \citet{wu2020dmm} and \citet{diakonikolas2017sq}, numbered results and section references follow
arXiv:1807.07237v2 and arXiv:1611.03473v2, respectively.
\end{remark}

\begin{remark}[Literature context]\label{hn:rem-literature}
For the Gaussian-mixture order indexed by the number of redundant components, the values
$\rbar(1)=4$, $\rbar(2)=6$ and the bound $\rbar\ge7$ are restated in the subsequent literature.
The mixture-of-experts literature instead indexes by the number $m$ of fitted experts in a cell:
\citet{nguyen2023softmax} establish $\rbar_{\mathrm{SM}}(2)=4$ and report
$\rbar_{\mathrm{SM}}(3)=6$ with the defective witness described in
Remark~\ref{hn:rem-errata}; Proposition~\ref{hn:prop-witness} supplies a valid witness for the
latter value. \citet[Section~3, paragraph following equation~(7)]{nguyen2024topk} report those
numerical values and the prediction $\rbar_{\mathrm{SM}}(m)=2m$. In their formulation,
equation~(7) uses $|\alpha_2|+\alpha_3+\alpha_4=\eta_2$, whereas the later derivation in
equations~(16)--(17) uses $|\alpha_2|+\alpha_3+2\alpha_4=\eta_2$. The standard-softmax
system considered here uses the weight-two constraint. \citet{hai2025dendrogram} also record that prediction.
\citet[Section~3.4]{nguyen2026ot} state the values
$4$, $6$, and at least $7$ for one, two, and at least three redundant components, respectively,
and say that the optimal uniform minimax rate remains unknown.
\end{remark}
%
%
%
%
%
%

%
%
%
%
%

%
%
\newif\ifskeleton\skeletonfalse   %
\ifskeleton\nocite{*}\fi
\clearpage
\section{Numerical methods and reproducibility}\label{app:note-experiments}

This appendix specifies the population-density calculations in
Section~\ref{sec:note-experiments}. The accompanying code and data archive contains the
scripts and raw outputs for these calculations.
The stored tables retain all density-evaluation points, including the 85 constructive-path
comparisons and their roots, moment-matching errors, signed mass integrals and transport distances,
as well as all fixed-size configurations used in the summaries.

\subsection{Distances, scaling and arithmetic}

We use
\[
\TV(p,q)=\frac12\int_{\RR}|p-q|\,\dd x,
\qquad
\Hel(p,q)^2=\frac12\int_{\RR}(\sqrt p-\sqrt q)^2\,\dd x,
\]
and $\chisq(p\|q)=\int(p-q)^2/q$.
The tables report $\Hel$ and $\sqrt{\chisq}$, so a distance order $2K$
corresponds to squared-divergence order $4K$.
For $G=\sum_jp_j\delta_{(\theta_j,v_j)}$ and $G_0=\delta_{(0,1)}$,
\[
W_1^{\mathrm E}(G,G_0)
 =\sum_jp_j\sqrt{\theta_j^2+(v_j-1)^2},
\qquad
\|G\|_{\mathrm h}^2
 =\sum_jp_j\sqrt{\theta_j^4+(v_j-1)^2}.
\]
The homogeneous parameterisation uses $\theta_j=\ell a_j$ and
$v_j=1+2\ell^2b_j$, with
$\sum_jp_j\sqrt{a_j^4+4b_j^2}=1$.
For the homogeneous-size experiments $\ell=\varepsilon$; for the
$W_1^{\mathrm E}$ experiments a scalar root solve fixes that distance
to $\varepsilon$.

The calculations used Python~3.11.5 with NumPy, SciPy, mpmath and Matplotlib. OMP, OpenBLAS and MKL thread counts were
set to one; independent tasks ran in separate CPU processes.

\subsection{High-precision constructive paths}

We construct the standard-normal quadrature rule through the symmetric
Jacobi matrix with off-diagonal entries $\sqrt j$, using 110-digit
mpmath arithmetic. The component means are $t x_j$, the variances are
$1-t^2$, and the quadrature weights are fixed. For every $K=2,\ldots,6$,
the 17 values of $t$ are
\[
\begin{gathered}
0.5,\ 0.35,\ 0.25,\ 0.18,\ 0.125,\ 0.09,\ 0.0625,\ 0.045,\ 0.03125,\\
0.0221,\ 0.015625,\ 0.011,\ 0.0078125,\ 0.0055,\ 0.00390625,\ 0.002,\ 0.001.
\end{gathered}
\]
TV is obtained by integrating the signed density difference between
numerically located sign changes. A scan of the scaled difference on
$[-14,14]$ uses 2,800 cells, followed by bracketed root refinement;
the integrations extend to the infinite tails. Hellinger integration uses
\[
\Hel(p,q)^2=\frac12\int_{\RR}
 \frac{(p-q)^2}{(\sqrt p+\sqrt q)^2}\,\dd x,
\]
with fixed integration breakpoints at $-8,-4,-2,0,2,4,8$ and the infinite
tails. This form avoids subtracting two nearly equal square roots.

For the reported leading-term comparisons, the TV normaliser is
$\kappa_K t^{2K}$ from \cref{hn:lem-gh-tv}; the stored Hellinger
normaliser is $|d_K|\sqrt{(2K)!/8}\,t^{2K}$, with
$d_K=-K!/(2K)!$. Table~\ref{tab:note-gh-endpoint} gives the endpoint
ratios.

\begin{table}[htbp]
\centering
\caption{Recorded leading-term ratios on the constructive path at
$t=10^{-3}$. The final column is the numerically located sign-change
count. Values are shown at the reported precision.}
\label{tab:note-gh-endpoint}
\begin{tabular}{@{}rrrr@{}}
\toprule
$K$ & TV / leading term & $\Hel$ / leading term & Sign changes\\
\midrule
2 & 1.00000106666759 & 1.00000000000257 & 4\\
3 & 1.00000160714464 & 1.00000000000201 & 6\\
4 & 1.00000213333623 & 1.00000000000320 & 8\\
5 & 1.00000265151941 & 1.00000000000464 & 10\\
6 & 1.00000316484104 & 1.00000000000633 & 12\\
\bottomrule
\end{tabular}
\end{table}

\subsection{Coefficient-system search and baseline checks}

To test coefficient cancellation numerically, we compute the coefficients recursively for
each component:
\[
e_0=1,\quad e_1=a,\quad
e_\alpha=\frac{a e_{\alpha-1}+2b e_{\alpha-2}}{\alpha},
\qquad E_\alpha=\sum_jp_je_\alpha(a_j,b_j).
\]
At homogeneous size one, the objective is
$\sum_{\alpha=1}^R\alpha!E_\alpha^2$, with
$R=2K-1,2K,2K+1$. Positive weights are parameterised by softmax.
Each available family supplies 400 starts: random, perturbed quadrature,
near-equal variance, merging pairs, tiny weight, widely separated
locations, with real projections of complex solutions added for $K=3,4,5$. Thus $K=2$ uses
2,400 starts per truncation and $K=3,4,5$ use 2,800.
The real least-squares solver uses the trust-region reflective method,
tolerances $10^{-15}$ and at most 4,000 function evaluations per start.

As baseline checks, the solver is also run on the order-$(2K-1)$ real system and on the
order-$2K$ complex system using 100 complex starts. For $K=3,4,5$, the complex solver reaches
objective values below $10^{-20}$ in 9, 22 and 56 starts, respectively; the smallest recorded
objective for $K=2$ is $2.37493146\times10^{-9}$.
Table~\ref{tab:note-coefficient-objectives} reports the real search's best objective values.

\begin{table}[htbp]
\centering
\caption{Smallest coefficient objective found at homogeneous size one.
These are sums of squared weighted coefficients.}
\label{tab:note-coefficient-objectives}
\begin{tabular}{@{}rrrr@{}}
\toprule
$K$ & $R=2K-1$ & $R=2K$ & $R=2K+1$\\
\midrule
2 & $8.38082876\,10^{-35}$ & $0.0296423083$ & $0.0328742658$\\
3 & $1.09048260\,10^{-33}$ & $0.000940991237$ & $0.000940991237$\\
4 & $9.27154712\,10^{-33}$ & $0.0000576511687$ & $5.77030278\,10^{-5}$\\
5 & $8.86527277\,10^{-33}$ & $0.00000217377594$ & $0.00000217377594$\\
6 & $2.19605855\,10^{-31}$ & $9.05174075\,10^{-8}$ & $9.05174075\,10^{-8}$\\
\bottomrule
\end{tabular}
\end{table}

The supplementary constrained search uses SLSQP to move the coefficient
$E_{2K}$ along solutions of the preceding equations for $K=2,\ldots,5$.
The fixed-size search reuses these solutions as initial configurations.

\subsection{Fixed-size density searches}

At each of eight geometrically spaced sizes from 0.4 to 0.03 for
$K=2,3,4$, or from 0.4 to 0.05 for $K=5$, the first search optimises
70 Hermite coefficients. Physical variances lie in $[0.1,1.9]$.
The objective is evaluated by Levenberg--Marquardt, with tolerances
$10^{-15}$ and at most 3,000 function evaluations per start.
Starts include the quadrature configuration, retained low-objective
coefficient configurations, up to ten constrained configurations,
15 random starts, seven quadrature perturbations and continuation.
Each saved best configuration is re-evaluated with a 160-term coefficient
series at 60 decimal digits. This supplies the recorded
$\sqrt{\chisq}$ values used for the log--log slope over the four smallest
sizes.

We then independently integrate TV and H for the saved configurations, using
$40+3K$ decimal digits.
The TV routine scans $[-12,12]$ in 2,400 cells to locate sign changes,
then integrates between the refined roots and over the infinite tails.
H uses the stable formula above.

For the direct Hellinger search, we use the same size and variance constraints for
$K=2,3,4$. Its double-precision
objective uses 1,281 points on $[-16,16]$ with step 0.025.
Each size receives 48 random starts, 24 perturbed quadrature starts,
the quadrature configuration, eight saved same-mode coefficient-search
configurations and one order-$2K$ configuration, followed by a
continuation start after the first size, giving 82 or 83 starts. Levenberg--Marquardt uses tolerances $10^{-15}$ and at most
4,000 function evaluations. Each winner is re-evaluated by high-precision
TV/H integration; the resulting H values supply the reported slopes.

\begin{table}[htbp]
\centering
\caption{Stored fixed-size slopes over the four smallest sizes. Columns
give the coefficient-search slope of $\sqrt{\chisq}$ and the direct-H
search slope after high-precision integration. The $W_1^{\mathrm E}$ rows
show the finite-scale effect of changing the size constraint.}
\label{tab:note-size-comparison}
\begin{tabular}{@{}lrrr@{}}
\toprule
Size constraint & $K$ & $\sqrt{\chisq}$ search & Direct H search\\
\midrule
$\|G\|_{\mathrm h}$ & 2 & 4.00002624 & 4.00006882\\
 & 3 & 6.00001999 & 6.00002040\\
 & 4 & 8.00002396 & 7.99999148\\
 & 5 & 9.99949221 & ---\\
\midrule
$W_1^{\mathrm E}$ & 2 & 3.98713528 & 3.98721777\\
 & 3 & 5.84036305 & 5.84036830\\
 & 4 & 7.95296041 & 7.95293527\\
 & 5 & 9.65255665 & ---\\
\bottomrule
\end{tabular}
\end{table}

For the smallest homogeneous-size $K=4$ direct-H configuration,
the grid objective gives $3.17446504\times10^{-15}$ and the
high-precision integral gives $3.18071148\times10^{-15}$.
The reported slope uses the high-precision evaluations.

\end{document}